\documentclass[11pt]{article}

\usepackage[preprint,nonatbib]{neurips_2021}

\usepackage[utf8]{inputenc} 
\usepackage[T1]{fontenc}    
\usepackage{hyperref}       
\usepackage{url}            
\usepackage{booktabs}       
\usepackage{amsfonts}       
\usepackage{nicefrac}       
\usepackage{microtype}      
\usepackage{xcolor}         
\usepackage{amsmath}
\usepackage{graphicx}
\usepackage{cite}
\usepackage[square,numbers,sort&compress]{natbib}
\title{Can sparse autoencoders make sense of gene expression\\latent variable models?}

\author{%
  Viktoria Schuster$^{1,2}$\\
  $^1$ Eric and Wendy Schmidt Center,\\
  Broad Institute of MIT and Harvard\\
  $^2$ Department of Computer Science,\\
  University of Copenhagen\\
  \texttt{vschuster@broadinstitute.org}\\
}

\begin{document}

\maketitle

\begin{abstract}
  Sparse autoencoders (SAEs) have lately been used to uncover interpretable latent features in large language models. By projecting dense embeddings into a much higher-dimensional and sparse space, learned features become disentangled and easier to interpret. This work explores the potential of SAEs for decomposing embeddings in complex and high-dimensional biological data. Using simulated data, it outlines the efficacy, hyperparameter landscape, and limitations of SAEs when it comes to extracting ground truth generative variables from latent space. The application to embeddings from pretrained single-cell models shows that SAEs can find and steer key biological processes and even uncover subtle biological signals that might otherwise be missed. This work further introduces \textbf{scFeatureLens}, an automated interpretability approach for linking SAE features and biological concepts from gene sets to enable large-scale analysis and hypothesis generation in single-cell gene expression models.
\end{abstract}

\section{Introduction}

Neural networks have proven to be powerful tools for analyzing complex data, yet they often lack inherent interpretability from a human perspective \citep{rudin_stop_2019}. While various approaches like disentanglement \citep{rauker_toward_2023}, adversarial training \citep{rauker_toward_2023}, and over-determined networks \citep{radhakrishnan_wide_2023} have shown some success in improving model interpretability \citep{marcinkevics_interpretability_2023,rauker_toward_2023}, they fall short of providing a comprehensive understanding of all learned features within a model \citep{rudin_interpretable_2022}. Recent research has revealed that features in neural networks are often learned in a state of superposition \citep{elhage_toy_2022}, where individual neurons encode multiple features (termed polysemanticity), and single features are distributed across multiple neurons. Simply speaking, each feature superposition is a linear combination of all dimensions in the latent space. In light of this complexity, sparse autoencoders (SAEs) \citep{olshausen_sparse_1997} have emerged as a promising tool for interpreting entire neural network layers \citep{sharkey_interim_2022,bricken_towards_2023,huben_sparse_2023,gao_scaling_2024}. The application of SAEs to large language model layers has demonstrated remarkable success in reducing polysemanticity, effectively translating language model activations into singular, monosemantic features \citep{sharkey_interim_2022,bricken_towards_2023,huben_sparse_2023,gao_scaling_2024}. However, this research has primarily been limited to language models and transformer architectures. Given that superpositions are strongly influenced by data structure \citep{elhage_toy_2022}, there is a pressing need to extend this approach to different types of hidden streams and data domains.
\\
Biology and health present a wealth of complex data and machine learning applications \citep{senders_machine_2018,lima_deep_2021,zhang_graph-based_2022,corti_artificial_2023,pun_ai-powered_2023,habineza_end--end_2023}. Single-cell gene expression (scRNAseq) data, for example, provide valuable insight into cellular functions and malfunctions within the human body. However, the high dimensionality and noise inherent in this data present significant analytical challenges \citep{kharchenko_triumphs_2021,lahnemann_eleven_2020,heumos_best_2023}. 
Several generative models have been suggested to model scRNAseq and multi-omics data and produce lower-dimensional representations for analysis \citep{heumos_best_2023,argelaguet_computational_2021,xu_probabilistic_2021,lopez_deep_2018,ashuach_multivi_2023,lin_scjoint_2022,stark_scim_2020,yang_multi-domain_2021,zuo_deep-joint-learning_2021,zuo_deep_2021,minoura_mixture--experts_2021}. Representation learning is of high interest in this field, as it is generally assumed that these high-dimensional biological processes are guided by lower-dimensional concepts such as regulatory programs.
\\
This work investigates the limitations and potential applications of SAEs for high-dimensional and sparse single-cell gene expression data. It examines superpositions and SAE features derived from models trained on simulated data and applies SAEs to pre-trained models \citep{schuster_multidgd_2023,geneformer}. Code for reproducibility is \href{https://github.com/viktoriaschuster/interpreting_omics_models}{available here}. The core insights and contributions are:
\vspace{-0.2em}
\begin{itemize}
    \itemsep0em
    \item Distribution type and distance of hidden generative variables affect variable recovery.
    \item SAEs extract meaningful features from single-cell expression models that successfully steer cells into desired programs. Features can act either locally or globally.
    \item \textbf{scFeatureLens}: An analysis pipeline for interpreting single-cell expression models by automatically annotating SAE features with biological concepts derived from ontologies, \href{https://github.com/viktoriaschuster/sc_mechinterp}{available on GitHub}.
\end{itemize}
\vspace{-0.5em}

\section{Related work}

The application of SAEs and dictionary learning in general has attracted a lot of attention in the field of natural language processing \citep{sharkey_interim_2022,bricken_towards_2023,huben_sparse_2023,gao_scaling_2024}. Recent research has demonstrated the efficacy of these methods in uncovering fine-grained features within language models, such as identifying hierarchical semantic structures \citep{yun_transformer_2021}, specific scriptures \citep{bricken_towards_2023}, and causal features of object identification \citep{huben_sparse_2023}. Others have presented improvements in the tradeoff between sparsity and reconstruction, reduced the occurrence of dead neurons, and developed metrics for evaluating quality based on hypothesized features \citep{gao_scaling_2024}. While much of the focus has been on language models, efforts to enhance interpretability have extended to other architectural domains. \citet{bau_network_2017} developed a method for scoring convolutional activations based on pre-defined visual concepts, thereby enhancing our understanding of learned visual features. 
\\
In contrast to these advancements, the application of SAEs to the field of biology has been limited. Except for recent applications to protein language models \cite{simon_interplm_2024,adams_mechanistic_2025}, dictionary learning has primarily been employed as a direct method for learning sparser representations \citep{rams_dictionary_2022,lopez_learning_2023,hao_dictionary_2024} or aligning representations more closely with specific biological concepts such as pathways \citep{karagiannaki_learning_2023}. More commonly, efforts to enhance the interpretability of biological representations have focused on disentanglement. Disentanglement is often applied to separate technical bias from biological signal through approaches such as adversarial training \citep{guo_integration_2022}, sparsity-inducing priors \citep{lopez_learning_2023}, overcomplete autoencoders \citep{zhang_graph-based_2022}, or architectural modularity \citep{piran_disentanglement_2024,schuster_multidgd_2023}.

\section{Sparse autoencoders}\label{sec:vanilla}

In representation learning, data is generally assumed to exist on a lower-dimensional manifold due to dependencies between features \citep{bengio_representation_2013}. Reducing the dimensionality into a latent representation through unsupervised learning can help reveal underlying structure. 
With a different constraint than dimensionality, data structure can also be revealed in a higher-dimensional setting by employing sparsity constraints on the latent representation \citep{olshausen_sparse_1997}. This has lately been exploited to disentangle the polysemanticity of hidden layers in large language models \citep{sharkey_interim_2022,bricken_towards_2023,huben_sparse_2023,gao_scaling_2024}. Figure \ref{fig:sim_superposition}A shows a schematic of SAEs and superpositions.
\\
\textbf{Vanilla SAE:} The simplest SAE maps an input $\mathbf{x} \in \mathbb{R}^d$ to a higher-dimensional hidden activation vector $\mathbf{z} \in \mathbb{R}^{l}_{\geq 0}$ and back, with an additional objective to promote sparsity in the activation space. The encoder is defined as 
\begin{equation}
    \mathbf{z} = \mathrm{ReLU} (\mathbf{W}_{\phi}(\mathbf{x}) + \mathbf{b}_{\phi})
\end{equation}
and the decoder as
\begin{equation}
    \mathbf{\hat{x}} = \mathbf{W}_{\theta}(\mathbf{z}) + \mathbf{b}_{\theta}
\end{equation}
with $\phi$ and $\theta$ indicating encoder and decoder parameter sets, respectively. The loss is given by 
\begin{equation}
    \mathcal{L} = \| \mathbf{x} - \hat{\mathbf{x}} \|^2_2 + \lambda \|\mathbf{z}\|_1
\end{equation}
where the first term is the mean squared error (MSE) loss for reconstruction. The second term is the sparsity penalty in the form of an $\mathrm{L1}$ loss weighed by hyperparameter $\lambda$, which will be referred to as the $\mathrm{L1}$ weight.
\\
\textbf{Other SAE setups:} A widely used version of the SAE uses an additional pre-network bias $\mathbf{b}_{pre}$ term applied to $\mathbf{x}$ before encoding \citep{bricken_towards_2023}, which has shown to improve performance \citep{elhage_toy_2022}. 
$k$-sparse autoencoders additionally use a different activation function (TopK) to directly control the number of active neurons (removing the need for the $\mathrm{L1}$ loss) \citep{makhzani_k-sparse_2014}. 
The latest advance in SAE research has been to reduce the number of dead hidden neurons by initializing encoder $\mathbf{W}_{\phi}$ and decoder $\mathbf{W}_{\theta}$ as transposes of each other and including dead neurons in an auxiliary loss \citep{gao_scaling_2024}.

\section{Simulation experiments}

As recent use cases of SAEs are mainly limited to the activations of large language models, this work presents an analysis of some common SAEs in a simulated setting with known underlying variables. The simulated data are inspired by sparse count data as we see in (single-cell) expression. 
Two datasets were created, a ``small'' one for hyperparameter sweeps with lower dimensionalities and a ``large'' one with realistic number of samples and dimensions in the observed variables $Y$ (the ``counts''). The simulation is based on a hierarchical generative process starting with hidden variables $X$ representing core programs, cell-type specific factors $A$, and batch effects $B$ with defined connectivity $\mathbf{M}$ of shape $(|Y|,|X|)$. The underlying hypotheses data simulation process are explained in detail in Appendix \ref{sec:exp_simulation} and depicted in Figure \ref{fig:sim_superposition}B. What follows is a discussion of what aspects of the data generation process can be recovered in superposition and SAE features, as well as performance differences of ``Vanilla'', ``ReLU'' \citep{bricken_towards_2023}, and ``TopK'' \citep{gao_scaling_2024} SAE architectures. 

\subsection{What is learned in superposition?}\label{sec:simsec1_superposition}

\textbf{Experimental set up:} Autoencoders were trained with a variety of structures and training hyperparameters (Appendix \ref{sec:sim_ae_arc}, Table S\ref{tab:hyperparams_ae}) on observables $Y$ of the large simulation data. 
Learned representations were extracted and used to compute superposition vectors and fits through linear regression (Appendix \ref{sec:meth_superposition}).

\begin{figure}[h]
\centering
\includegraphics[width=1\linewidth]{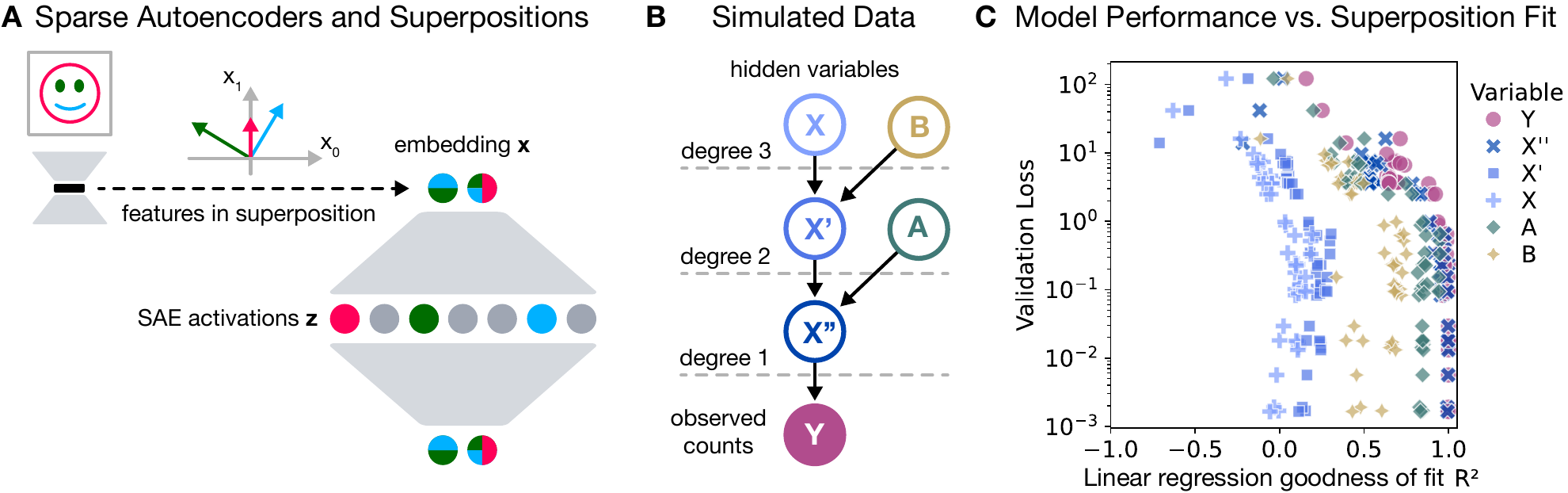}
\caption{\textbf{Sparse Autoencoders, data simulation, and hidden variable recovery. A} Schematic of superpositions and SAEs. Given a sample generated from 3 features and encoded into a 2D latent space, there are more features than dimensions. The features have to be learned as linear combinations of the latent dimensions (meaning they are in superposition). These features can be disentangled by projecting them into a higher-dimensional space via SAEs. \textbf{B} Schematic of the data generation process. Filled and non-filled circles represent observed and hidden variables, respectively. Arrows indicate the dependencies between variables from parent to child. There are 3 levels to this generative process, indicated by the distance from observed counts $Y$ (details in Appendix \ref{sec:exp_simulation_large}). $X'$ and $X''$ represent the hidden states altered by $B$ and $A$, respectively. \textbf{C} AE performance (validation loss) plotted against superposition fit ($R^2$). Coefficients of determination $R^2$ describe how well a given variable can be retrieved from the latent embedding (see Appendix \ref{sec:meth_superposition} for details). Colors and marker styles match the variables from B.}
\label{fig:sim_superposition}
\end{figure}

\textbf{Results:} Observables are perfectly learned when validation loss is sufficiently low, and hidden variables can be partially recovered from latent representations (Figure \ref{fig:sim_superposition}C). Recovery of variables follows a distinct pattern: variables $X''$ directly upstream of $Y$ are most accurately reconstructed, followed by $A$, $B$, $X'$, and $X$. Regression fits $R^2$ did not scale linearly with the distance from $Y$, suggesting that the type of variables and their role in the data generation process influence recovery. Additionally, recovery of more distant variables seemed to decrease with larger (deep and wide) models despite lower validation loss (Figure S\ref{fig:supp_sim_super}). 

\subsection{How do different SAEs perform?}

\textbf{Experimental set up:} A sweep of different SAE architectures and a wide range of hyperparameters (Table S\ref{tab:hyperparams_sae}) was performed on embeddings from an autoencoder trained to perfectly recover observables and hidden variables $X$ of the small simulation data (Appendix \ref{sec:sim_ae_arc}). 
All SAEs were trained on the extracted representations and evaluation metrics were computed as described in Appendix \ref{sec:hyper_eval}.

\textbf{Results:} \textbf{Reconstruction and sparsity.} Briefly summarized, reconstruction losses of Vanilla and ReLU SAEs were more robust compared to TopK (Figures S\ref{fig:sim_metrics}A-B, S\ref{fig:supp_sae_performance}). 
Sparsity (fraction of dead/active neurons) strongly increased for $\mathrm{L1}$ weights above $10^{-3}$ and a $k$ below $50$\,\% (Figure S\ref{fig:sim_scaling}), and strongly depended on the learning rate (Figure S\ref{fig:supp_sae_lr}). As a result, the analysis was continued with the overall best-performing learning rate of $10^{-4}$. 
\\
\textbf{Recovery of $X'$.} Figure S\ref{fig:sim_metrics} shows that a small hidden size can be detrimental to the performance and interpretability of TopK models. In terms of good recovery (high correlation between features and observables) with little redundancy for variables $X$ and $Y$, the Vanilla SAEs showed the best tradeoff and TopK the worst (Figures S\ref{fig:supp_sae_recovery_x},\ref{fig:supp_sae_recovery_y}). 
The best performing Vanilla models used $\mathrm{L1}$ weights of $10^{-3}$ ($10^{-4}$ for ReLU) with hidden dimensionalities of $5-50\times$ the size of the latent space (for best recovery and 1-5 neurons per variable). 
The number of features per variable scaled roughly exponentially for the $k$-sparse autoencoder (TopK) over the hidden dimension irrespective of $k$ (Figure S\ref{fig:sim_scaling}). For Vanilla and ReLU SAEs, there was no such scaling tendency and the $\mathrm{L1}$ weight strongly determined the rate at which the number of neurons per variable grow, which is a disadvantage of these SAEs.

\subsection{How well can data variables and structure be recovered?}

\textbf{Experimental set up:} SAEs were trained on AE embeddings of the large simulation data from section \ref{sec:simsec1_superposition} according to the results from the previous sweep (Appendix \ref{sec:sim_structure}). They were evaluated in terms of correlation between SAE neuron activations and data variables, and to what extent the structure of the generative connectivity matrix $\mathbf{M}$ is recovered by the SAE. Cosine similarities between observables and SAE neurons $(|Y|,|z|)$ were used to create pseudo connectivity matrices for different thresholds. These pseudo connectivity matrices were compared to $\mathbf{M}$ through Binomial tests (Appendix \ref{sec:sim_structure}).

\textbf{Results:} \textbf{Variables.} Recovery of a given variable from SAE features was measured as the correlation between that variable and SAE neuron activations. 
Observed variables $Y$ and directly upstream hidden variables $X''$ could be nearly perfectly recovered, especially for larger hidden dimensionalities. The original generative random variables $X$, however, are not directly represented by individual SAE features. 
Comparing these results to baselines from PCA, ICA, and SVD (Table S\ref{tab:recovery_baseline}), there was no significant increase in superposition identifiability between SAE and baseline methods. The SAE's advantages, however, lie in the discovery of unknown features and providing a convenient way of extracting learned features that can be used for model steering.
\\
\textbf{Structure.} In real-world applications, it may be difficult to identify generative variables due to the prevalence of features corresponding to observables. We may, however, be able to identify concepts and structures in the data generation process in a different way. Figure S\ref{fig:sim_recovery_a} demonstrates an alternative approach comparing the structure of SAE features and observables with the data generation matrix $\mathbf{M}$. 
For each feature, the best matching $X''$ variable is determined. 
Their entries of pseudo connectivity matrix and $\mathbf{M}$ are used to calculate how many entries of $Y$ match for each feature-$X''$ pair. On average, $75\,\%$ to $95\,\%$ of the entries in $\mathbf{M}$ could be recovered (with $20$th to $70$th percentiles of cosine similarity as thresholds, respectively).

\section{Case Study: Extracting and annotating meaningful features from single-cell models}

Next, SAEs were applied to representations from models pre-trained on single-cell RNAseq and multi-omics data. SAE hyperparameters were evaluated on a model trained on three different datasets from \citet{schuster_multidgd_2023}. Meaningfulness of extracted features and how they can be used for steering samples towards biological programs is demonstrated in a manual evaluation. 
A major contribution of this work is an automated analysis pipeline for practical large-scale interpretability analysis demonstrated on multiDGD \citep{schuster_multidgd_2023} and the latest version of Geneformer \citep{geneformer}. 
In this case study, Gene Ontology (GO) terms \citep{ashburner_gene_2000,the_gene_ontology_consortium_gene_2023}, which provide functional information about sets of genes, represent examples of biological concepts.

\subsection{SAE training}

\textbf{Experimental set up:} SAE hyperparameters were evaluated on a small sweep for extracted representations from multiDGD instances trained on human bone marrow \citep{lueckenSandboxPredictionIntegration}, mouse gastrulation \citep{argelaguet_decoding_2022}, and human brain data \citep{trevino_chromatin_2021} (Appendix \ref{sec:sc_data}). 
Results scaled well compared to the preceding simulation experiments. Hyperparameters and training are described in Appendix \ref{sec:sc_sae_train}. The final SAE hidden dimension was chosen to be $10000$ neurons in favor of redundant features over a lack of sensitivity. 
Another SAE was trained on representations of the human bone marrow data extracted from Geneformer for the automated pipeline. 
See Appendices \ref{sec:sc_data} and \ref{sec:sc_sae_train} for embedding extraction, training details and compute estimates.

\textbf{Results:} In the SAE trained on multiDGD embeddings from the human bone marrow data, $5318$ remained as ``live'' SAE neurons with $185.7$ firing on average per cell. Since the representations are highly structured with respect to cell type
, average activations of cell types naturally create unique patterns (Figure S\ref{fig:supp_sc_activ}). Significant differences in activations with respect to cell types revealed two major SAE feature categories: ``local'' and ``global'' (categorization and significance measure in Appendix \ref{sec:signif}). Local features are characterized by higher activations for a single cell type compared to all other cell types. 
Among the $5318$ live neurons, there were $4410$ global and $908$ local features. Training the SAE on different random seeds revealed robust results in the number of live neurons and feature types (Table S\ref{tab:sc_sae_randomseed}). 
Monocytes and cells along the red blood cell differentiation trajectory accounted for most of the local features (not related to numbers of cells in the data, Figure S\ref{fig:supp_sc_local}). 

\begin{figure}[h]
\centering
\includegraphics[width=1\linewidth]{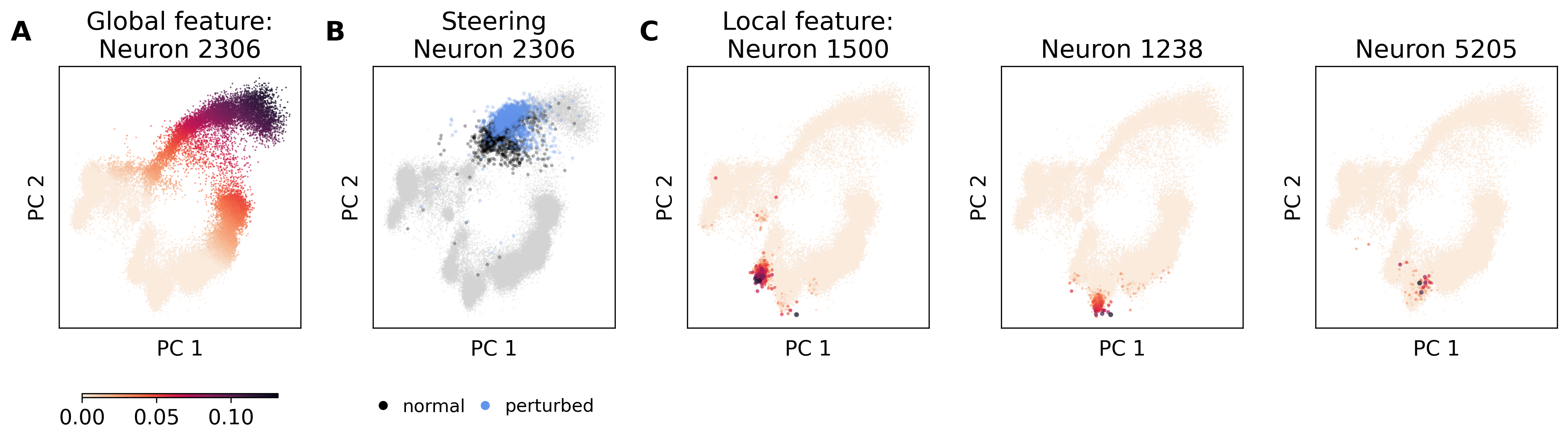}
\caption{\textbf{SAE features in multiDGD' bone marrow representation space.} Visualized as PCA plots of the extracted 20-dimensional representations. \textbf{A} Representations colored by activations of neuron $2306$. Lightest color represents zero values, darker colors present higher activations. \textbf{B} Representations from SAE feature steering/perturbation experiments on Proerythroblast representations (black, ``normal''). Representations predicted by the SAE after maximizing feature $2306$ are shown in blue (``perturbed''). \textbf{C} Local features. From left to right: Representations colored and size-scaled by activations of neurons $1238$, $5205$, and $1500$.}
\label{fig:sc_reps}
\end{figure}

\subsection{Manual feature analysis}

\textbf{Experimental set up:} Evaluating what biological potential functions a feature has is difficult. In this work, concepts of biological function of a given feature was approximated by GO terms. In order to create gene sets associated with a given SAE feature, Differential Gene Expression (DGE) analysis was performed on either ``perturbed-vs-normal'' or ``high-vs-low'' sample subsets. Perturbed subsets were created by selecting a cell type along the global feature trajectory, computing sample activations, maximizing the feature of interest (also called ``steering''), and predicting the perturbed representations. ``High-vs-low'' subsets were created by selecting the $95$th and $5$th percentile activations of sample representations per feature (excluding 0 if done in a specific cell type). 
DGE analysis was then performed based on the single-cell model's predicted expression values according to Appendix \ref{sec:diff_expression}.

\textbf{Results:} \textbf{Global features.} Red blood cell differentiation is a prominent biological process in this dataset. Based on the rule set described in Appendix \ref{sec:exp_rules}, neuron $2306$ was identified as the best aligning feature (Figure S\ref{fig:supp_sc_perturb}). Activations are shown in Figure \ref{fig:sc_reps}A. 
Although feature $2306$ was most prevalent along the axis of red blood cell differentiation, moderate activations were also found in NK and some CD8+ T cells. 
Steering was performed by maximizing feature $2306$ in HSCs, Proerythroblasts, NK, and CD8+ T cells (Figure \ref{fig:sc_reps}B). 
While each analysis resulted in different gene sets and GO terms, the identified processes are highly specific and show a strong functional overlap (Table S\ref{tab:go}). Results highlight ion homeostasis and gas transport, which are crucial processes in erythropoiesis and cytotoxicity. This global feature presents an important higher-level and more general concept in cellular processes of the bone marrow.
\\
\textbf{Local features.} Among local features, B cells presented multiple of the top $20$ features regarding mean activation. 
This analysis investigates one of the most significant local features for each of the three different types of B cells present in the data: Transitional, Naive CD20+, and B1 B cells. Activations are shown in Figure \ref{fig:sc_reps}C. DGE analysis (``high-vs-low'') and GO term enrichment analysis within each cell type revealed distinctive molecular signatures of each feature. \underline{Feature 1500 (Transitional B cells)} was characterized by GO terms related to the response to interferon beta. Interferon beta is a critical regulator during early transitional B cell development, playing a role in differentiation towards a regulatory phenotype vs. an inflammatory phenotype \citep{schubert_interferon-_2015}. \underline{Feature 1238 (Naive CD20+ B cells)} showed enrichment in histone H3R26 citrullination, an indicator of cellular aging \citep{zhu_histone_2021}. Another sign of cell aging is increased closed chromatin. Cells with high activations of feature $1238$ had significantly more closed chromatin. The $95$th percentile had an average chromatin openness of $0.03322 \pm 0.00124$ SEM compared to the $5$th percentile with a mean of $0.04393 \pm 0.00002$ (based on ``high'': $35$ samples, ``low'': $3483$ samples, $129921$ columns). \underline{Feature 5205 (B1 B cells)} presented enriched GO terms predominantly centered around molecular functions associated with pattern recognition receptor activities. Specifically, the terms highlighted activation of the innate immune system, referencing key receptors such as toll-like receptor 4, haptoglobin, and RAGE receptor. The activation profile of these cells suggests a trajectory towards increased immune cell activity and potential cytotoxicity, paralleling observations from previous results on T cells.

\begin{figure}[h]
\centering
\includegraphics[width=1\linewidth]{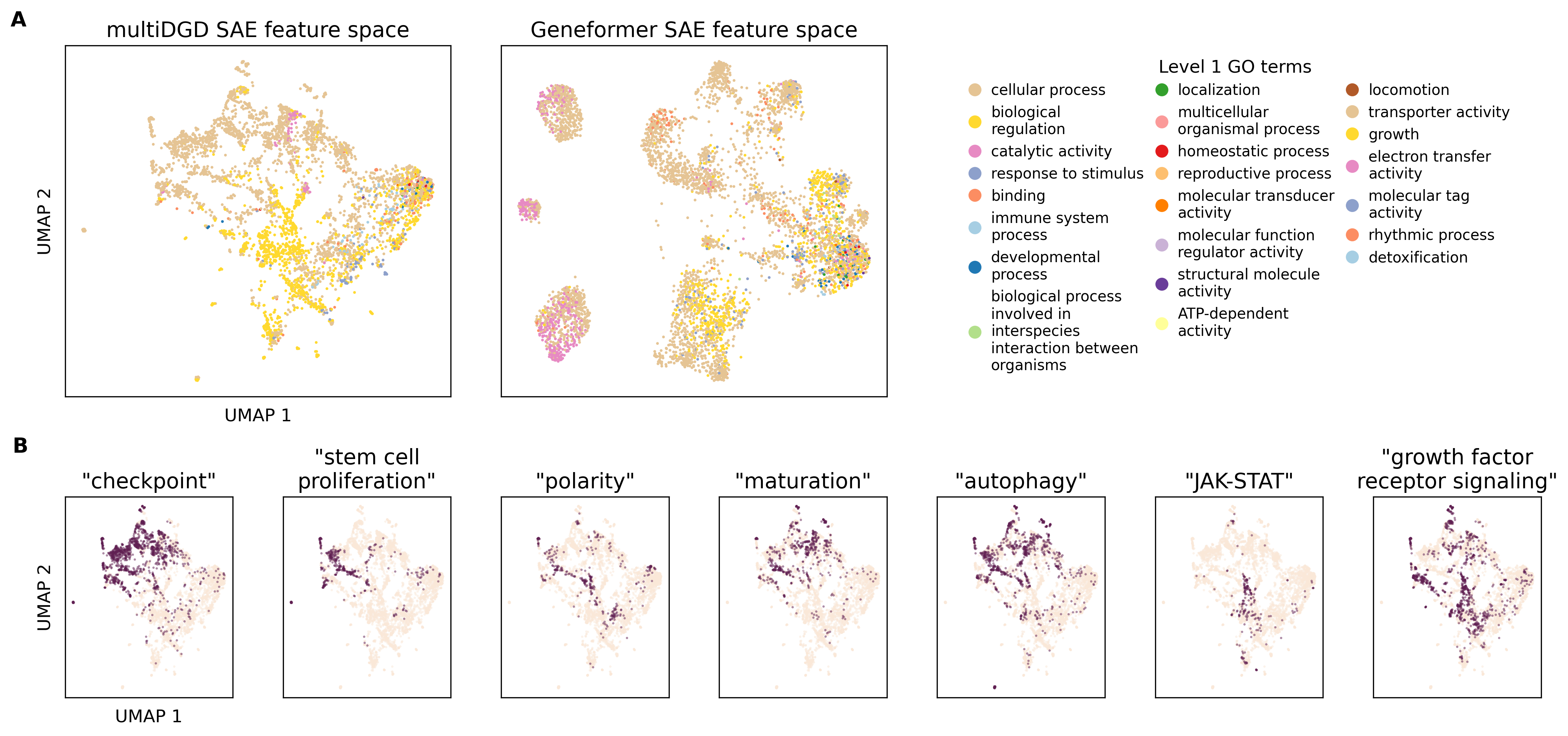}
\caption{\textbf{SAE feature space.} Feature spaces are visualized as UMAPs of the GO-feature matrices described in Appendix \ref{sec:automatic_analysis}. Locations of the manually analyzed features are shown in Figure S\ref{fig:sim_feat_location}. \textbf{A} Feature maps of SAEs from multiDGD (left) and Geneformer (right) representations are colored by the most common 1st level of GO terms associated with the feature (legend on the right). \textbf{B} Probing multiDGD SAE features by broad semantic concepts (plot titles) included in GO terms. Dark points indicate features with at least one GO term containing the concept.}
\label{fig:sc_automatic}
\end{figure}

\subsection{scFeatureLens: Automated SAE analysis demonstrated on multiDGD and Geneformer}
Manual analyses, while useful for validation, are limited in their scalability. Deriving biological semantic concepts in an automated fashion is highly desirable and a key contribution of this work. 
The pipeline presented here can be adapted to any database using gene sets to characterize semantic concepts. The automated analysis in this work is performed both on the previously introduced SAE features trained on multiDGD representations of human bonemarrow data and an equivalent SAE trained on Geneformer \citep{geneformer} representations from the same data.

\textbf{Experimental set up:} The basis of the automated analysis is a concept-by-gene matrix summarizing the gene sets associated with each concept. ``high-vs-low'' sample sets were created for each active feature with the $99$th percentile of the feature activations as the ``high'' set and a sample of maximum $1000$ cells from those with zero values as ``low''. This was followed by DGE analysis on the predicted expression counts and a simple GO term analysis inspired by \citet{mi_large-scale_2013} (details in Appendix \ref{sec:automatic_analysis}). The analysis is parallelized over GO terms for efficiency with a compute time of $\sim 30$ seconds per feature on the used Hardware. GO terms with p-values below $0.01$ were recorded for each feature. Feature spaces are visualized as UMAPs \citep{mcinnes_umap_2020} of the binary matrix of the matches between unique GO terms and features (distance 1.0, 10 neighbors, seed 0, spread 10).

\textbf{Results:} The analysis on multiDGD's SAE returned GO terms for $4374$ ($82.25\,\%$) of the active features, with overall $1875$ unique biological process and $624$ molecular function GO terms. 
Individual GO terms appeared between once and over $2500$ times. Terms that appeared very often are broader, high-level GO terms associated with immune response and signaling pathways (Table S\ref{tab:top_go}). 
Many of the features active in a small fraction of cells did not cluster with cell types and would go completely unnoticed in traditional analysis of the dense latent space, making this pipeline very valuable. Figures \ref{fig:sc_automatic}A and S\ref{fig:dgd_umap} show the SAE features' concept space. 
This space organizes features with respect to GO terms, largely separating into cellular processes and biological regulation at the highest level of the Gene Ontology. It can be probed for specific biological components and concepts, which is demonstrated in Figures \ref{fig:sc_automatic}B and S\ref{fig:sae_probing_supp}. Examples of meaningful overlaps include a large overlap of features associated with protein localization and checkpoint signaling. Within this area there are processes that collectively contribute to stem cell homeostasis, fate determination and maintenance, such as stem cell proliferation, cell polarity, maturation, and autophagy. The JAK-STAT signaling pathway takes a central role in this feature space. It appears at the intersection of features annotated with concepts from growth factor signaling, NK cell activation, antiviral response, and death - to name a few key functions. 
\\
The SAE trained on Geneformer representations resulted in $7073$ active features ($33\,\%$ more than the SAE trained on multiDGD) out of which $5290$ were annotated with GO terms. Interestingly, there are only $409$ local features. This potential lack of local separability may be due to the curse of dimensionality (Geneformer has a latent dimensionality of $896$ vs multiDGD's $20$) and the complex latent distribution of multiDGD. See visualizations of embedding and more feature space plots in Figures S\ref{fig:sae_geneformer_embedding}-\ref{fig:geneformer_umap}. Feature spaces are difficult to compare. Visually, many observations made previously in terms of overlapping concepts seem to be consistent, although more specific GO terms do not cluster well (Figure S\ref{fig:sae_geneformer_probing}). Additionally, all $2499$ unique GO terms identified in the multiDGD SAE were also recovered from the Geneformer embeddings, with $97$ additional GO terms found in this larger, pre-trained model. The most common GO terms center less around immunity, which is not very suprising given that Geneformer was trained on a large and more varied dataset (Table S\ref{tab:top_go}). Concept count distributions between the two models' SAEs varied with Spearman and Pearson correlations of $0.46$ and $0.43$, respectively (Figures S\ref{fig:go_counts_geneformer},\ref{fig:sae_geneformer_dgd_go_counts}). multiDGD's SAE shows an average of $95.5 \pm 1.7$ SEM GO terms per active feature with a range from $1-482$. The SAE trained on Geneformer embeddings has a lower range of GO terms per feature from $1-254$ with an average of $48.7 \pm 1.0$ SEM. A more fine-grained analysis of the feature spaces through optimal bipartite matching based on the shared GO terms reveals a low similarity of 0.16 (Appendix \ref{sec:matrix_comarison} for methodology and interpretation). These results suggest that there may be a shared broad semantic structure that is learned by different models on the same kind of data, but individual features seem to potentially serve very different purposes and the functional focus of the embeddings are largely influenced by the scope of data the model was trained on.

\section{Conclusion}

This work explored the potential of sparse autoencoders (SAEs) to interpret latent representations in biological tabular data. Through data simulation with ground-truth generative variables, it provided valuable insights into the behavior and capabilities of SAE architectures. SAEs were found to effectively recover hidden variables if they have been learned in superposition, with performance improving as hidden dimensionality and model width increase. 
The presence of hidden variables in superposition depends, however, on their position in the data generation process, the impact they have on the observables, and likely also their type of distribution. Variables with an indirect effect on the observed data and little structure in the generative process could practically not be recovered. SAEs further do not pose an advantage in the recovery of known or hypothesized features compared to simple baselines. However, the connectivity of SAE features and observables can unearth valuable insight into the data generation structure. 
\\
Despite their limitations, the application of SAEs to single-cell expression data demonstrated that they present practical value in a real-world biological context. 
Identifying and steering features manually uncovered specific biological processes, validating the relevance of the SAE-derived features. Local features helped identify small cell type subpopulations previously not distinguishable in the latent representations. The automated annotation pipeline employs well-established methods such as DGE and enrichment analysis. Its novelty and utility stem from direct integration with the disentangled SAE features extracted from scRNAseq embeddings. This provides a novel, powerful, and scalable framework for 
improved interpretability. It is available as a tool \href{https://github.com/viktoriaschuster/sc_mechinterp}{on GitHub}. While this case study was limited to Gene Ontology (GO) terms which are incomplete and biased towards well-studied genes, the improvement in interpretability is immense and can have a significant impact on single-cell analysis. Additionally, the pipeline can be applied to any gene expression embedding and can be used with different databases providing semantic context from gene sets. 
\\
Altogether, this work presents an important step towards more interpretable models in biology, but much more research is needed in this field. Future work could explore metrics for evaluating the biological meaningfulness in and differences between embeddings, and methods to help overcome the limitations in recovering variables that are difficult to decompose.

\bibliography{references}


\section*{Code availability}

Code is made available for reproducibility in \href{https://github.com/viktoriaschuster/interpreting_omics_models}{this GitHub repository}. The presented tool \textbf{scFeatureLens} is available in \href{https://github.com/viktoriaschuster/sc_mechinterp}{this GitHub collection}.

\section*{Data availability}

All data and models used in this work are publicly available and cited.

\section*{Conflict of Interest}

I declare no conflict of interest.

\section*{Acknowledgements}

I would like to acknowledge great discussions with coworkers at the Center of Health Data Science (University of Copenhagen) and the Eric and Wendy Schmidt Center at the Broad Institute. I thank the reviewers for their time and effort to help improve the quality and communication of the work. I especially want to thank my mentor Anders Krogh for his tremendous support. I also want to thank Uthsav Chitra and Kristoffer Stensbo-Smidt for their feedback and advice, and Jonas Sindlinger for being a wonderful rubber duck.

\section*{Impact Statement}

This paper presents work whose goal it is to advance the development and application of mechanistic interpretability for the fields of biology and medicine. There are many potential positive impacts for society related to improving disease understanding and treatment. A potential negative impact with interpretability of biological models is the exploitation of knowledge about differences related to gender, ethnicity, socioeconomic background, and genetics. I believe, however, that the open source development of interpretability techniques will lead to both discovery and removal of such biases in biological models.

\appendix

\clearpage

\onecolumn
\renewcommand{\figurename}{Supplementary Figure}
\renewcommand{\tablename}{Supplementary Table}
\setcounter{figure}{0}
\setcounter{table}{0}
\setcounter{section}{1}

\section*{Appendix A: Methods}

\subsection{Compute infrastructure}

All computations were performed using \texttt{Python 3.9} on either CPU or one of the following GPUs: NVIDIA A30, NVIDIA RTX A5000.

\subsection{Simulated Data}

\subsubsection{Data Simulation}\label{sec:exp_simulation}\label{sec:exp_simulation_large}\label{sec:small_sim}

\begin{figure*}[h]
\centering
\includegraphics[width=1\linewidth]{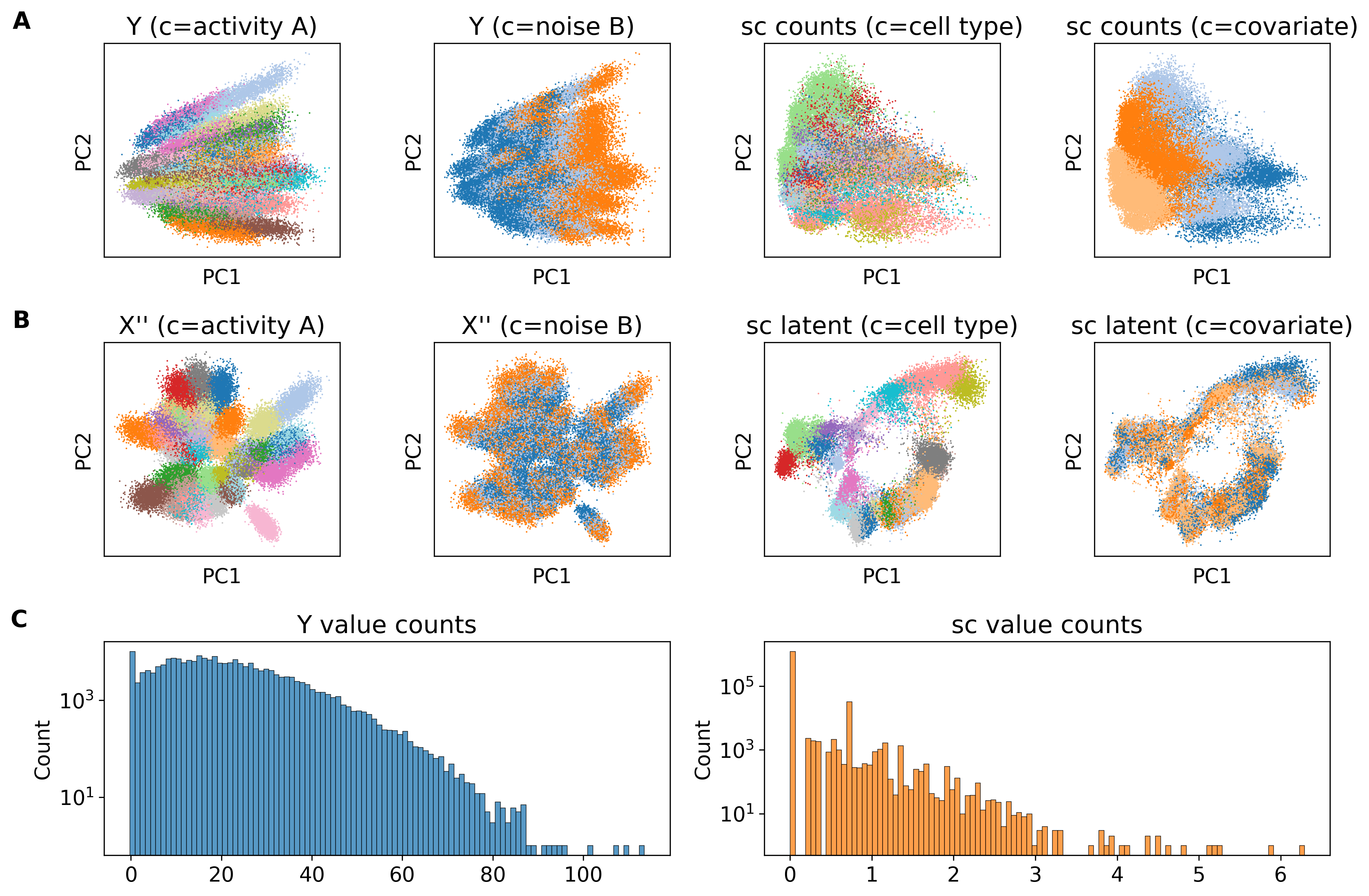}
\caption{\textbf{Simulated and single-cell data.} \textbf{A} PCAs of simulated observables $Y$ and log-transformed single-cell (sc) counts colored by $A$/celltype and $B$/technical covariate, respectively. \textbf{B} PCAs of simulated latents $X''$ and inferred (not generative) latents from the sc model. PCAs are again colored by $A$/celltype and $B$/technical covariate, respectively. \textbf{C} Histograms of simulated $Y$ values and real sc counts. Simulated data does not directly match the specific single-cell dataset presented here. However, clusters of $A$ and $B$ appear similar to our real-world comparison (cell type and technical covariate). The values in C are generally higher for the simulation and less sparse, but still match zero-inflated Negative Binomial distributions which are typically used to describe these count data.}
\label{fig:supp_sim_data}
\end{figure*}

Simulated data sets were designed with inspiration from sparse count data as we see in single-cell sequencing in order to get an understanding of what SAEs learn about the data structure and hidden variables. First, a set of hypotheses is defined to guide our the generation process:
\begin{itemize}
    \itemsep0em
    \item Gene regulation is determined by molecular regulators and gene programs, and thus the observed data $\mathcal{Y}$ should lie on a lower-dimensional manifold $\mathcal{X}$.
    \item Different cell types $L$ have different patterns of active regulators/programs and different levels of overall expression $\mathcal{Y}$.
    \item Technical noise or other covariates $\mathcal{B}$ can cause shifts in $\mathcal{Y}$.
\end{itemize}
Simulated counts $\mathbf{y} \in \mathcal{Y} = \{Y_{i=1}, ..., Y_{i=N}\}^{\mathrm{T}}$ are generated through the following three steps:
\begin{equation}\label{eq:large_sim}
\begin{split}
    \mathbf{x'} &= \mathbf{x} + \mathbf{b_c} \text{ with } \mathbf{x} \sim \mathcal{X}\\
    \mathbf{x''} &= \mathbf{x'} \, \mathbf{a_{c}}\\
    \mathbf{y} &= \sum_{j=1}^{100} m_{j} \, \mathbf{x''_{j}}\\
\end{split}
\end{equation}
with $\mathcal{X} = (X_1, ..., X_{100})^{\mathrm{T}}$ presenting the ground truth multivariate latent variables. Noise vectors $\mathbf{b_c} = \mathcal{B}\mathbf{s_{c1}}$ and cell type activity vectors $\mathbf{a_c} = \mathbf{A}^{\mathrm{T}}\mathbf{s_{c2}}$ are products of one-hot selection column vectors $\mathbf{s_c}$ with noise distribution $\mathcal{B} = (B_1, ..., B_3)^{\mathrm{T}}$ and activity matrix $\mathbf{A} = (\mathrm{a}_{lj}) \in \mathbb{N}_0^{40\times100}$, respectively. Matrix $\mathbf{M} = (\mathrm{m}_{ij}) \in \mathbb{N}_0^{N\times100}$ presents the connectivity matrix between regulators/programs and genes. Random variables were sampled according to 
\begin{equation}
\begin{split}
    X_j &\sim \mathrm{Pois}(\lambda=1.1j)\\
    B_g &\sim \mathcal{N}(\mu=j,\sigma=0.1)\\
    \mathbf{s_{c1}} &\sim \mathrm{Cat}(p=\frac{1}{3},k=3)\\
    \mathrm{a}_{lj} &\sim \mathrm{Bin}(k=1,p=0.3)\\
    \mathbf{s_{c2}} &\sim \mathrm{Cat}(p=\frac{1}{40},k=40)\\
    \mathrm{m}_{ij} &\sim \mathrm{Bin}(k=1,p=0.1).
\end{split}
\end{equation}
For a ``large'' simulation with realistic dimensions, a data dimensionality of $N=20000$ was chosen which is at the upper limit of the number of protein-coding genes in the human genome \citep{amaral_status_2023}. The latent dimensionality of $\mathcal{X}$ was set to $100$. $L=40$ dimensions for $\mathbf{A}$ represent different cell types and $G=3$ variables in $\mathcal{B}$ simulate technical noise. Distribution parameters and the order of the generative process were chosen so that the simulated data $\mathcal{Y}$ would present similar structures and count distributions compared to real data (Supplementary Figure \ref{fig:supp_sim_data}). $90000$ train and $10000$ validation data points were sampled. For simplicity, all of the variables of interest will be referred to as $Y$ ($\{\mathcal{Y},\mathbf{y}\}$), $X$ ($\{\mathcal{X},\mathbf{x}\}$), $X'$ ($\mathbf{x'}$), $X''$ ($\mathbf{x''}$), $A$ ($\mathbf{a_c}$), $B$ ($\mathbf{b_c}$). Additionally, a ``small'' simulation set was created for a large-scale SAE sweep and the possibility to visually verify superpositions. It features $|Y|=5$, $|X|=3$, $L=1$, and no noise. Details can be found in Appendix \ref{sec:small_sim}.

The small simulation data set with $|Y|=5$ and $|X|=3$ was generated in two steps. First, the three-dimensional multivariate random variable $X$ were sampled from Binomial distributions with probabilities $[0.5, 0.1, 0.9]$ multiplied with samples from Poisson variable $A$ ($\lambda=2$), resulting in latent variables $X'$. Secondly, $X'$ was multiplied with $\mathbf{M}$ ($p=0.1$) to produce observables $Y$. $10000$ train and $2000$ validation data points were sampled.
\begin{equation}\label{eq:small_sim}
\begin{split}
    \mathbf{x'} &= \mathbf{x} \, \mathbf{a} \text{ with } \mathbf{x} \sim \mathcal{X}\\
    \mathbf{y_i} &= \sum_{j=1}^{3} m_{i,j} \, \mathbf{x'_{j}}
\end{split}
\end{equation}

\subsubsection{AE architectures and training}\label{sec:sim_ae_arc}
\begin{figure*}[h]
\centering
\includegraphics[width=1\linewidth]{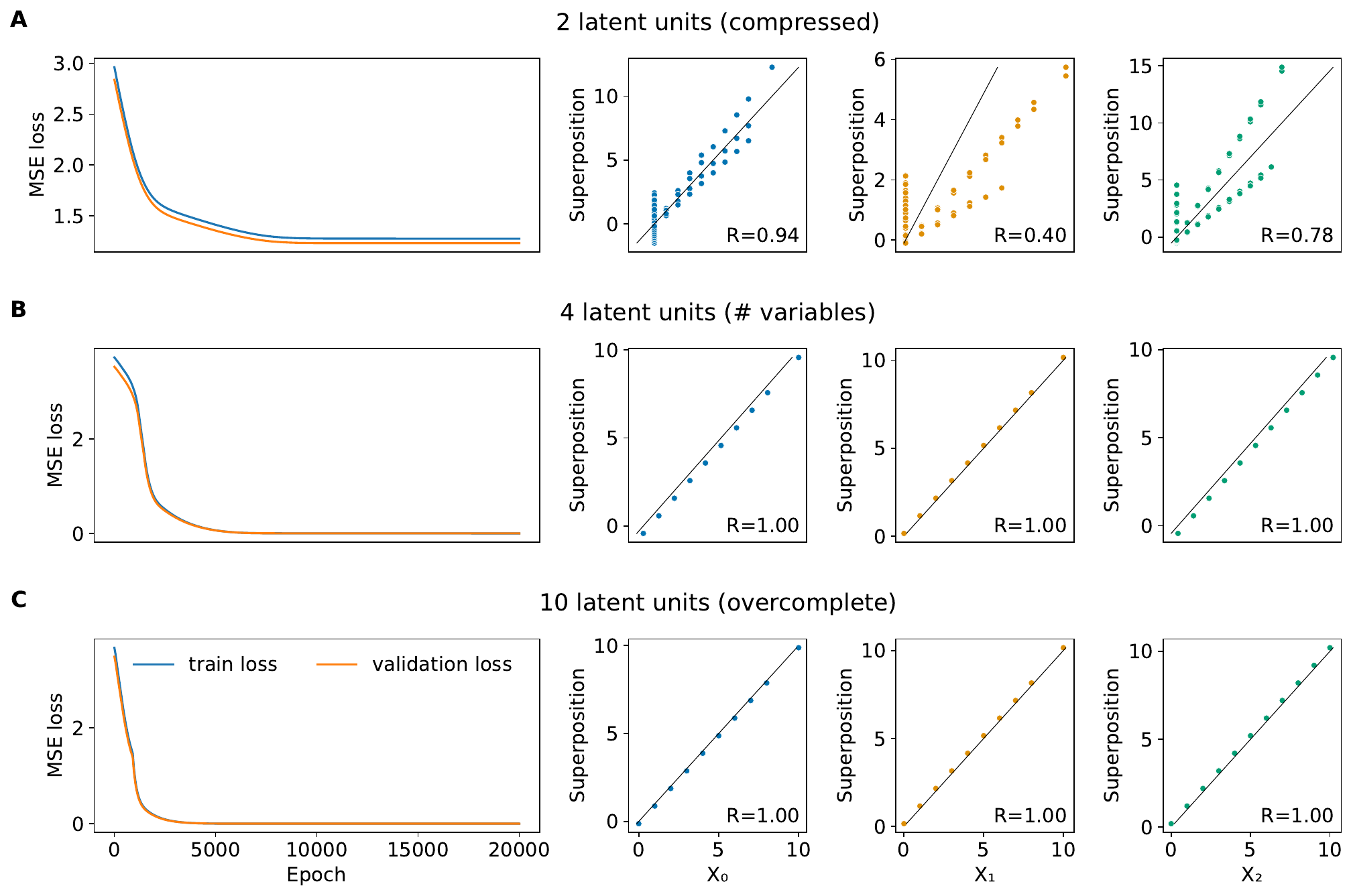}
\caption{\textbf{Superpositions in compressed, ``ideal'', and overcomplete autoencoders trained on simulated data. A) } The top row depicts learning curves of train and validation MSE loss over epochs (left, legend in C) and superpositions of the three variables $X$ (right) of a single-layer autoencoder with a compressed bottleneck (2 dimensions). The superpositions are plotted as the product of the latent representations and coefficients from linear regression against the true values of $X$. Linear regression was performed between the latent representations and true $X$ values. Points along the black line indicate a perfect fit of the superpositions (quantified by the R value rounded to two decimals in the bottom right corner (maximum 1). \textbf{B)} Same as A for the ``ideal'' case, in which the number of latent units is equal to the number of generative random variables. \textbf{C)} Same as A and B for the overcomplete case with 10 hidden units.}
\label{fig:supp_sim_ae_comp}
\end{figure*}

An autoencoder was trained that perfectly recovered latent variables $X'$ (Supplementary Figure \ref{fig:supp_sim_ae_comp}) of the small simulation data with latent dimension $4$ equal to the number of generative variables, ReLU activation, Adam optimizer \citet{kingma2014method} (learning rate $10^{-4}$), and MSE loss for $20000$ epochs. 
\\
Autoencoder architectures for the large simulation were set up as either ``narrow'' or ``wide'' with mirrored encoder and decoder. $d$ here is referred to as the latent dimensionality. A ``narrow'' encoder would be of structure $[\max(1000, 2d), \max(150, 2d), ... , \max(150, 2d)]$ unless the number of layers was only $2$, in which case the hidden dimensionality would be $\max(150, 2d)$. A ``wide'' encoder would receive hidden dimensionalities sampled from equidistant points between the input dimension and $d$. Hyperparameters were determined through Optuna optimization \citet{optuna_2019} based on the reconstruction loss with 50 trials and 100 epochs. The trials tested learning rates between $10^{-6}$ and $10^{-3}$, weight decays $[0, 0.1, ..., 10^{-7}]$, dropout $[0, 0.1]$ and batch sizes between $32$ and $512$. Selected hyperparameters for each depth and width can be found in Table S\ref{tab:ae_optuna}. Remaining parameters are shown in Table S\ref{tab:hyperparams_ae}. All models were trained with Adam optimizer and early stopping for up to $10000$ epochs. 

\subsubsection{Superpositions}\label{sec:meth_superposition}

Superpositions in latent representations were identified through linear regression. For the small simulations, superposition vectors and coefficients of determination ($R^2$) were computed through \texttt{sklearn}'s \texttt{LinearRegression}. For the sake of effiency on the large number of variables in the large simulation, linear regression was implemented using a single linear neural network layer trained for $100$ epochs by optimizing the mean squared error with standard gradient descent optimization and a learning rate of $10^{-4}$. 

Given observed values ${y_i}$ with mean $\bar{y}$ and predicted values ${\hat{y}_i}$, the coefficient of determination
\begin{equation}
    R^2 = 1 - \frac{\sum_i (y_i - \hat{y}_i)^2}{\sum_i (y_i - \bar{y}_i)^2}
\end{equation}
represents the fraction of variance explained by the model, with values ranging from 0 (no explanatory power) to 1 (perfect prediction).

\subsubsection{SAE hyperparameter evaluation}\label{sec:hyper_eval}
Different SAE architectures were trained on varying hidden dimensionalities (latent size multiplied with a hidden factor), learning rates, and $\mathrm{L1}$ weights for 500 epochs. All tested hyperparameters can be found in Table S\ref{tab:hyperparams_sae}. In the case of TopK SAEs, the sparsity is controlled by $k$, which was tested as percentages of the hidden dimension. For each instance, the following metrics were computed:
\begin{itemize}
\setlength\itemsep{0em}
    \item[--] number of active hidden neurons (activity determined by activations of $> 10^{-10}$)
    \item[--] number of redundant hidden neurons (neurons that fire with other neurons with a Pearson correlation $\geq 0.95$)
    \item[--] average number of neurons firing per sample
    \item[--] average number of neurons corresponding to a given data variable (determined by Pearson correlation $\geq 0.95$)
    \item[--] highest Pearson correlation between a neuron and a given data variable
\end{itemize}

\subsubsection{Structure identification}\label{sec:sim_structure}

This analysis was done on one of the well-performing SAEs trained on representations from one of the best performing AEs in terms of validation loss and variable recovery. The AE featured $2$ layers in the ``wide'' format with $5075$ hidden neurons and a latent dimension of $150$. The SAE featured a scaling factor of $100$, an $\mathrm{L1}$ weight of $0.001$, and a learning rate of $10^{-5}$. Cosine similarities between all $11849$ active SAE features and all $20000$ observables in $Y$ were computed. Based on different percentiles of the cosine similarity matrix (as thresholds), connectivity matrices were computed between SAE features and $Y$ and Binomial tests between all features and all variables in $X''$ w.r.t. $Y$ were performed. The ground truth connectivity matrix was given by the data generation matrix $\mathbf{M}$. The best matching $X''$ for each feature was computed based on the maximum number of hits. The reported result is the maximum fraction of ``genes'' $Y$ connected to $X''$ covered by the set of ``genes'' $Y$ connected to the SAE features.


\subsection{Single-cell case study}\label{sec:exp_hema}

\subsubsection{Single-cell representations}\label{sec:sc_data}

Representations were extracted from three pre-trained multiDGD models \citep{schuster_multidgd_2023} trained on single-cell multi-omics data from human bone marrow \citep{lueckenSandboxPredictionIntegration}, mouse gastrulation \citep{argelaguet_decoding_2022}, and human brain \citep{trevino_chromatin_2021}. The following table of dataset sizes was taken from \citet{schuster_multidgd_2023}. 

\begin{table}[h]
    \centering
    \caption{Summary of single-cell multi-omics data used.}
    \begin{tabular}{lllll}
    Dataset & Species & Number of cells & RNA dimensionality & ATAC dimensionality\\
    \hline
    Bone marrow & Human & 69249 & 13431 & 116490\\
    Brain & Human & 3534 & 15172 & 95677\\
    Gastrulation & Mouse & 56861 & 11792 & 69862\\
    \end{tabular}
\end{table}

The same train-validation-test splits were used as in \citet{schuster_multidgd_2023}. The latent space of the model is small with only $20$ dimensions. The paper highlighted the structure of the latent space, especially with regard to the clear trajectory of differentiation from stem cells to red blood cells (erythrocytes) \citep{schuster_multidgd_2023}. The pre-trained model and data were downloaded as instructed by \citet{schuster_multidgd_2023}. Furthermore, embeddings for the human bone marrow data were extracted from Geneformer \citep{geneformer} following \href{https://huggingface.co/ctheodoris/Geneformer/blob/main/examples/extract_and_plot_cell_embeddings.ipynb}{their instructions} to extract embeddings by passing the scRNAseq data through the most recent version of Geneformer ``gf-20L-95M-i4096''. The extracted embeddings had a dimensionality of $896$.

\subsubsection{Identifying a feature for red blood cell differentiation}\label{sec:exp_rules}
\textbf{Red blood cell differentiation:} This rule set was created to identify potential features of red blood cell differentiation:
\begin{enumerate}
\setlength\itemsep{0em}
    \item The average activation must be higher in the red blood cell line than in other cell types.
    \item Average activations must consistently increase from the stem cells to the final differentiation stage of red blood cells.
\end{enumerate}

\begin{figure*}[h]
\centering
\includegraphics[width=1\linewidth]{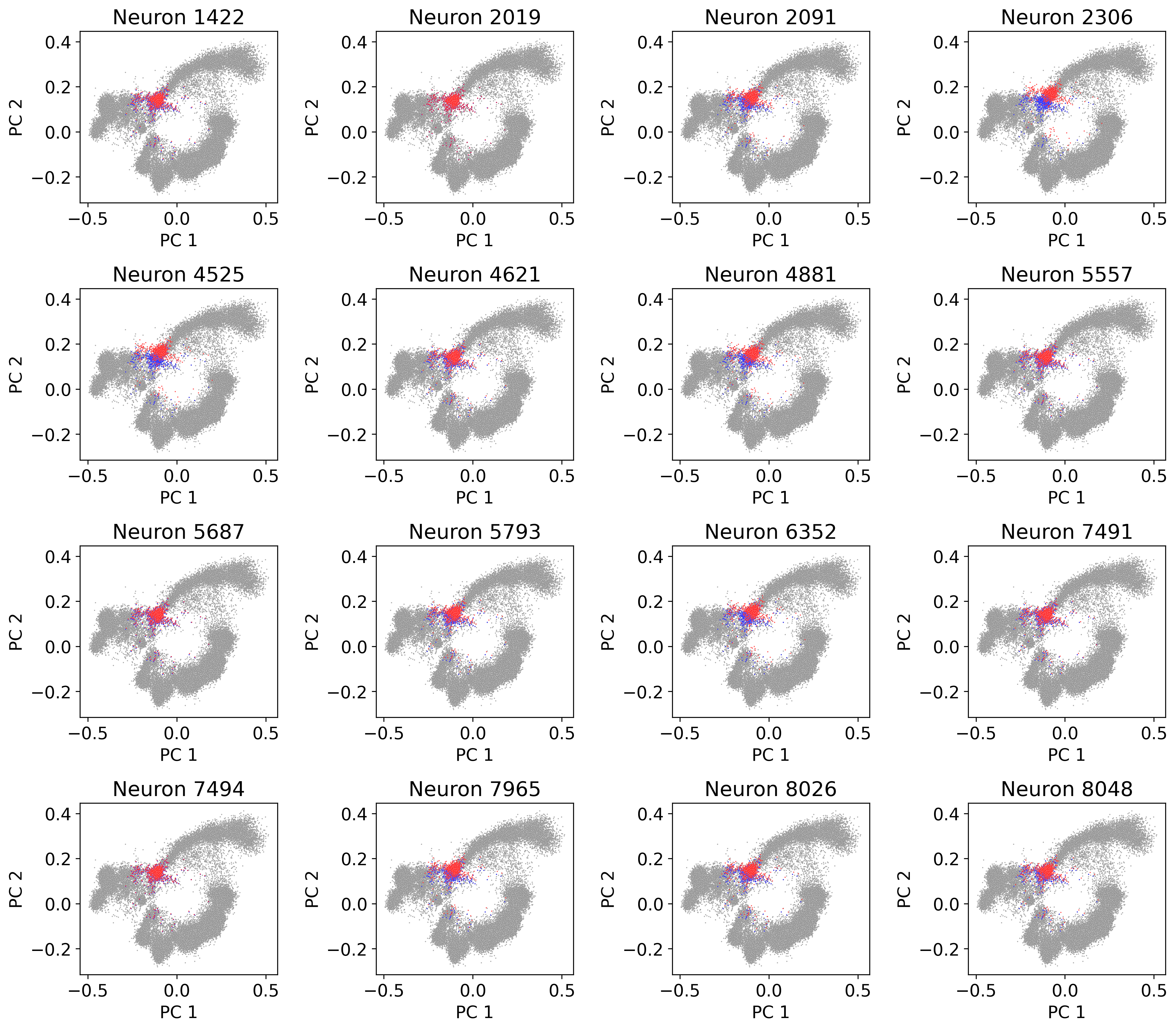}
\caption{\textbf{Effect of perturbations on potential features for red blood cell development.} PCA plots of the extracted single-cell representations (grey dots). Titles indicate the neuron that was perturbed. Blue and red dots present normal and perturbed samples, respectively.}
\label{fig:supp_sc_perturb}
\end{figure*}

Applying this rule set provided $44$ neurons as potential features. These neurons were inspected visually in terms of cell-wise activations and tested to see which ones would result in the largest shift in latent space towards differentiated cells when maximizing the neuron's activations in stem cells (Supplementary Figure \ref{fig:supp_sc_perturb}). See the next section for details on perturbations. This returned neuron $2306$ as the most promising candidate feature.

\subsubsection{SAE training}\label{sec:sc_sae_train}
A small hyperparameter search was performed on the multiDGD embeddings to see if the simulation results translated well to real world settings. Both Vanilla and Bricken SAEs were tested, but not TopK since this method was not robust in previous experiments and has the disadvantage of having to estimate the number of active neurons beforehand. Hyperparameters tested were hidden scaling factors $[20, 100, 200, 500]$, $L1$ weights $[1, 0.1, 0.01, 10^{-3}, 10^{-4}]$, and learning rates $[10^{-4}, 10^{-5}]$ with a batch size of $128$ for $1000$ epochs with early stopping (patience $50$).

A learning rate of $10^{-4}$ gave best results and was most robust, which aligns with simulation results. When training long enough, reconstruction loss generally decreased with the scaling factor. Learning rate $10^{-4}$ presented the lowest reconstruction losses and a much less drastic difference between scaling factors than smaller learning rates. Lower $L1$ weights lead to steeper increases in the number of active neurons against the scaling factor (again aligning with simulation results). Lower number of active neurons ($25th$ percentile) and good reconstruction loss ($5th$ percentile) can be achieved with learning rates of $10^{-4}$ and a $L1$ weights of $0.001$ or $0.0001$ (slight differences for datasets, shift by one log step). There were no trends or large differences between Vanilla and Bricken SAEs.

For analysis, Vanilla SAEs were trained for $500$ epochs with Adam optimizer \citet{kingma2014method}, a learning rate of $10^{-4}$, batch size $128$, hidden activation dimension $10000$ ($500$-fold increase for multiDGD) and an $\mathrm{L1}$ weight of $10^{-3}$ (see loss curves in Supplementary Figure \ref{fig:supp_sc_loss} for multiDGD human bone marrow) for multiDGD's and Geneformer's embeddings from the human bone marrow with random seeds $[0, 42, 9307]$. Compute requirements are low, with training taking $20$ minutes for the $56$k training samples.

\subsubsection{DGE analysis}\label{sec:diff_expression}
Sample groups were investigated in terms of relevant changes to gene expression through differential gene expression analysis (DGE). In the case of the ``perturbed-vs-normal'' paired samples, this was done with negative binomial generalized linear models as is common in biological data analysis \citep{anders_differential_2010,love_moderated_2014}. The resulting p-values and fold changes from the models are reported. For the unpaired ``high-vs-low'' comparison, t-tests were performed between the groups for each gene and calculated the fold change based on mean expression. Corrected p-values were computed based on multi-test correction with Benjamini/Hochberg correction for non-negative values \citep{benjamini_controlling_1995} for all experiments.

\subsubsection{Manual GO term enrichment analysis}
In order to identify biological processes related to the differentially expressed genes, genes were filtered by adjusted p-values (threshold $10^{-10}$) and in the case of \mbox{CD8+ T} cells also fold change (10-fold and inverse) to get as highly specific processes as possible. Biological processes related to the resulting gene sets were identified through GO term analysis with default parameters at \href{https://geneontology.org/docs/go-enrichment-analysis/}{https://geneontology.org/docs/go-enrichment-analysis/} \citep{ashburner_gene_2000,the_gene_ontology_consortium_gene_2023}.

\subsubsection{Feature characterization}\label{sec:signif}
SAE features were distinguished into local and global features based on whether they were only active in a single cell type or similarly active in multiple cell types. This was assessed by calculating the significance measures of activations per feature over cells from a specific cell type vs all other cells. Features with significantly higher activations in only one cell type were labeled as local. Significance was determined based on a two-tailed test with confidence interval $95\,\%$ ($z=1.96$) as
\begin{equation}
    \alpha = |\mu_j - \mu_i| - 1.96 (\frac{\sigma_j}{\sqrt{N_j}} + \frac{\sigma_i}{\sqrt{N_i}})
\end{equation}
with means $\mu$, standard deviations $\sigma$, and number of observations $N$ for two cell type distributions $i$ and $j$. The null hypothesis is rejected if $\alpha \geq 0.05$. This significance measure is used to determine relevant differences between samples for SAE feature activations and in one analysis also chromatin accessibility (openness).

\subsubsection{Automated GO term analysis}\label{sec:automatic_analysis}

DGE analysis was performed on the predicted expression counts for feature-specific ``high-vs-low'' sample sets as described in \ref{sec:diff_expression}. Next, a GO term analysis was performed according to \citet{mi_large-scale_2013} with a binomial test and a Mann-Whitney U (MWU) test for all GO terms with $20$ to $500$ reference genes available in our $13431$ genes. The MWU test were performed with the ranked fold changes (smallest rank 1). The metric was calculated as 
\begin{equation}
\begin{split}
    U = \min \, \Bigl( U_1 &= n_1 n_2 \frac{n_1 (n_1 + 1)}{2} - R_1,\\
    U_2 &= n_1 n_2 \frac{n_2 (n_2 + 1)}{2} - R_2 \Bigr)
\end{split}
\end{equation}
with $n_1$ and $n_2$ presenting the number of genes in the GO term gene set and the remaining genes, respectively. $R_1$ and $R_2$ correspondingly present the average ranks of these groups. $Z$-scores, p-values, and effect sized of the test are reported. The binomial test was conducted on the most relevant genes from the DGE analysis based on two thresholds. Firstly, the number of genes identified for an adjusted p-value threshold of $10^{-5}$ and a fold change of at least $2$ (or below $0.5$) were computed. If this returned zero genes, the p-value threshold was increased to $0.05$ and the fold change excluded. Afterward, the p-value, number of expected genes, fold enrichment and false discovery rate for $k$ hits (relevant genes that are also found in the GO term gene set), $n_s$ samples in the study (the relevant genes returned by DGE analysis), and $p_c$ as the probability of randomly finding one of the GO term genes ($p_c = n_c / n$ with $n$ as the total number of genes and $n_c$ as the number of genes associated with the GO term) were computed.

\subsubsection{Optimal bipartite matching}\label{sec:matrix_comarison}

Optimal bipartite matching computes the Jaccard distance between all features from DGD and Geneformer SAEs and then finds the optimal matching via the Hungarian algorithm. Overall matrix similarity is computed as the average Jaccard distance of the matched pairs. Values can be between 0 (no similarity) and 1 (perfect similarity).

\section*{Appendix B: Supplementary Materials}

\subsection*{Tables}

\begin{table*}[h]
\centering
\caption{Autoencoder hyperparameter configurations}
\label{tab:ae_optuna}
\begin{tabular}{lllll}
\textbf{Model Configuration} & \textbf{Dropout} & \textbf{Learning Rate} & \textbf{Weight Decay} & \textbf{Batch Size} \\
\hline \\
20-2-narrow   & 0.0   & 1e-4 & 1e-3 & 128 \\
20-2-wide     & 0.0   & 1e-5 & 1e-3 & 128 \\
20-4-narrow   & 0.0   & 1e-4 & 0.0  & 128 \\
20-4-wide     & 0.0   & 1e-5 & 1e-5 & 128 \\
20-6-narrow   & 0.0   & 1e-5 & 1e-5 & 128 \\
20-6-wide     & 0.0   & 1e-6 & 1e-5 & 512 \\
100-2-narrow  & 0.0   & 1e-4 & 1e-5 & 128 \\
100-2-wide    & 0.0   & 1e-5 & 1e-5 & 128 \\
100-4-narrow  & 0.0   & 1e-4 & 1e-7 & 128 \\
100-4-wide    & 0.0   & 1e-5 & 1e-5 & 128 \\
150-2-narrow  & 0.0   & 1e-4 & 1e-5 & 128 \\
150-2-wide    & 0.0   & 1e-5 & 1e-5 & 128 \\
150-4-narrow  & 0.0   & 1e-4 & 1e-7 & 128 \\
150-4-wide    & 0.0   & 1e-5 & 1e-5 & 128 \\
150-6-narrow  & 0.0   & 1e-4 & 1e-5 & 512 \\
150-6-wide    & 0.0   & 1e-5 & 0.1  & 256 \\
1000-2-narrow & 0.0   & 1e-5 & 1e-5 & 128 \\
1000-2-wide   & 0.0   & 1e-5 & 1e-5 & 128 \\
1000-4-narrow & 0.0   & 1e-4 & 0.0  & 128 \\
1000-4-wide   & 0.0   & 1e-5 & 1e-7 & 128 \\
1000-6-narrow & 0.0   & 1e-6 & 0.0  & 512 \\
1000-6-wide   & 0.0   & 1e-5 & 1e-7 & 512 \\
\end{tabular}
\end{table*}

\begin{table*}[h]
\caption{\textbf{Simulation autoencoder hyperparameters}} \label{tab:hyperparams_ae}
\begin{center}
\begin{tabular}{lll}
\textbf{Simulation} & \textbf{Hyperparameter} & \textbf{Choices}\\
\hline \\
small & latent dimension & $4$\\
 & learning rate & $10^{-4}$\\
 & Number of layers & 1 \\
 & Number of hidden neurons & n.a.\\
 & Batch size & 1\\
 & Weight decay & $0$ \\
 & Random seed & $0$\\
large & latent dimension & $[20, 100, 150, 1000]$\\
 & learning rate & see Table S\ref{tab:ae_optuna}\\
 & Number of layers & $[2,4,6]$\\
 & Number of hidden neurons & [``narrow'', ``wide''] (details in Appendix A \ref{sec:sim_ae_arc})\\
 & Early stopping & $20$ epochs\\
 & Batch size & see Table S\ref{tab:ae_optuna} \\
 & Weight decay & see Table S\ref{tab:ae_optuna} \\
 & Random seed & $[0,42,9307]$\\
\end{tabular}
\end{center}
\end{table*}

\begin{table*}[h]
\caption{\textbf{Simulation SAE hyperparameters}} \label{tab:hyperparams_sae}
\begin{center}
\begin{tabular}{lll}
\textbf{Simulation} & \textbf{Hyperparameter} & \textbf{Choices}\\
\hline \\
small & scaling factor & $[2,5,10,20,50,100,200,1000]$\\
 & learning rate & $[10^{-2},10^{-3},10^{-4},10^{-5}]$\\
 & $\mathrm{L1}$ weight & $[10^{-1},10^{-2},10^{-3},10^{-4},10^{-5}]$\\
 & k (TopK percent active neurons) & $[5, 10, 20, 50, 75, 100]$ \\
large & scaling factor & $[20,100,200,500]$\\
 & learning rate & $[10^{-4}, 10^{-5}, 10^{-6}]$\\
 & L1 weight & $[10^{-1}, 10^{-2}, 10^{-3}, 10^{-4}]$\\
 & Early stopping & 20 epochs\\
 & Batch size & $128$ \\
\end{tabular}
\end{center}
\end{table*}

\begin{table*}[h]
\caption{\textbf{Simulation variable recovery in an autoencoder with latent dimension 150.} Per hidden generative variable, the maximum Pearson correlation of all features against all variable dimensions are reported. For the SAE, an average of highest correlations over 4 SAEs with differen hidden scaling factors are reported $\pm$ SEM.} \label{tab:recovery_baseline}
\begin{center}
\begin{tabular}{llllll}
\textbf{Method} & $X''$ & $X'$ & $X$ & $A$ & $B$\\
\hline \\
PCA & \textbf{0.72} & 0.36 & 0.35 & -0.12 & 0.56\\
ICA & 0.54 & \textbf{0.50} & \textbf{0.50} & 0.02 & -0.01\\
SVD & 0.66 & 0.41 & 0.32 & -0.09 & 0.62\\
SAE & \textbf{0.74} $\pm$ 0.02 & 0.43 $\pm$ 0.01 & 0.19 $\pm$ 0.01 & -0.54 $\pm$ 0.01 & \textbf{0.65} $\pm$ 0.01\\
\end{tabular}
\end{center}
\end{table*}

\begin{table*}[h]
\caption{\textbf{Robustness of number of live neurons and feature types for different random seeds}} \label{tab:sc_sae_randomseed}
\begin{center}
\begin{tabular}{ll}
\textbf{Feature type} & \textbf{Mean $\pm$ SEM}\\
\hline \\
Local (live) & $914.00 \pm 2.45$\\
Global (live) & $4427.33 \pm 7.08$\\
Dead & $4658.67 \pm 9.53$\\
\end{tabular}
\end{center}
\end{table*}

\begin{table*}[h]
\caption{\textbf{Top 5 most abundant GO terms in the automated analysis}} \label{tab:top_go}
\begin{center}
\begin{tabular}{lll}
\textbf{GO name} (multiDGD)\\
\hline
immune response\\
cell surface receptor signaling pathway\\
structural constituent of ribosome\\
adaptive immune response\\
inflammatory response\\
\\
\textbf{GO name} (Geneformer)\\
\hline
cytoplasmic translation\\
translation\\
structural constituent of ribosome\\
chromatin binding\\
mRNA splicing, via spliceosome\\
\end{tabular}
\end{center}
\end{table*}

\begin{table*}[h]
\caption{\textbf{Feature 2306 Perturbation GO terms.} Go terms associated with the gene lists derived from DEG analysis for each cell type perturbation experiment. Only highly specific GO terms are shown with maximum 400 gene references. GO terms appearing for more than one experiment are highlighted in bold font. Abbreviations: CT - cell type, HSC - hematopoietic stem cell, PE - proerythroblast, NK - natural killer cell, CD8T - CD8+ T cell. } \label{tab:go}
\begin{center}
\begin{tabular}{lllll}
\textbf{GO term}  &\multicolumn{4}{l}{\textbf{Present in CT perturbation}} \\
 & HSC & PE & NK & CD8T\\
\hline \\
Intracellular \textbf{calcium} ion homeostasis & \checkmark & & & \\
\textbf{Carbon dioxide transport} & & \checkmark & \checkmark & \\
\textbf{Oxygen transport} & & \checkmark & \checkmark & \\
\textbf{Hydrogen peroxide catabolic process}  & & \checkmark & \checkmark & \\
Positive regulation of myoblast differentiation  & & \checkmark & & \\
Erythrocyte development  & & \checkmark & & \\
Vascular process in circulatory system  & & \checkmark & & \\
Nitric oxide transport  & & & \checkmark & \\
Stimulatory C-type lecitin receptor signaling pathway  & & & \checkmark & \\
Positive regulation of natural killer cell mediated cytotoxicity  & & & \checkmark & \\
Myeloid leukocyte activation  & & & \checkmark & \\
Chemokine-mediated signaling pathway & & & & \checkmark \\
\textbf{Calcium}-mediated signaling & & & & \checkmark \\
\end{tabular}
\end{center}
\end{table*}

\clearpage

\subsection*{Figures}


\begin{figure*}[h]
\centering
\includegraphics[width=1\linewidth]{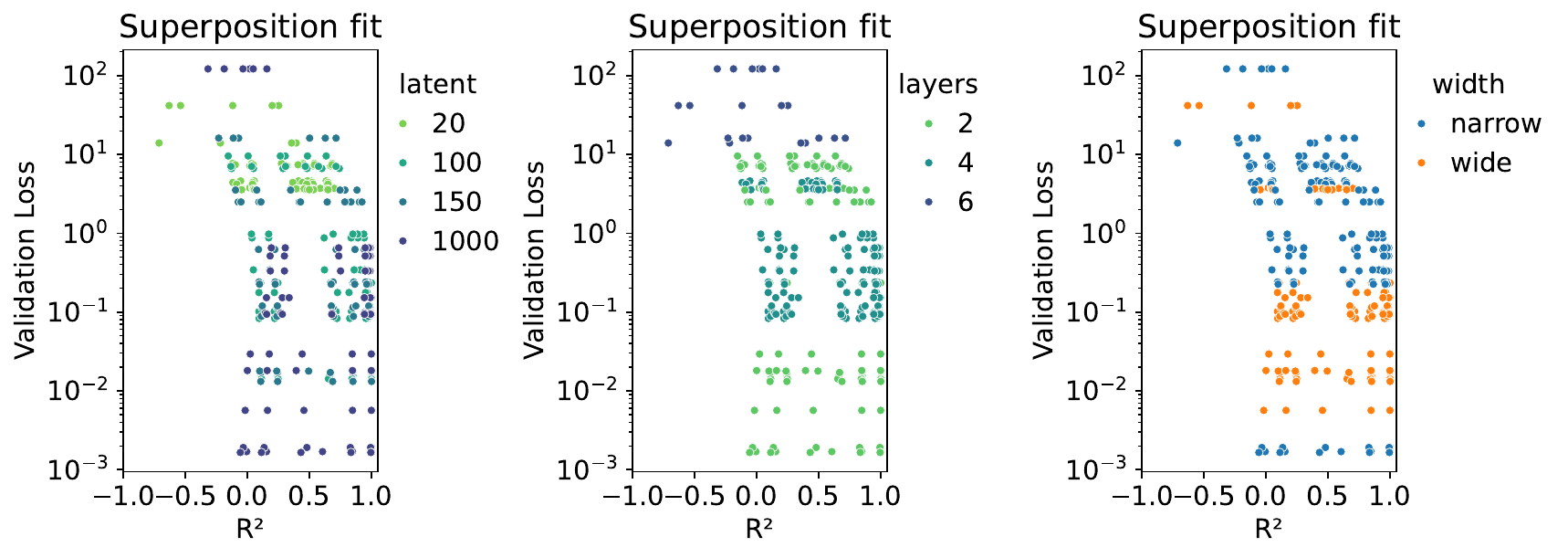}
\caption{\textbf{Validation loss against superposition fits for large simulation autoencoders.} AE performance (validation loss) vs. superposition fit. Coefficients of determination $R^2$ were computed based on linear regression performed on the AE latent representations w.r.t. each of the variables on the left. Colors present the latent dimension, number of hidden layers, and architecture width (details in Appendix A \ref{sec:sim_ae_arc}), respectively.}
\label{fig:supp_sim_super}
\end{figure*}

\begin{figure}[h]
\centering
\includegraphics[width=1\linewidth]{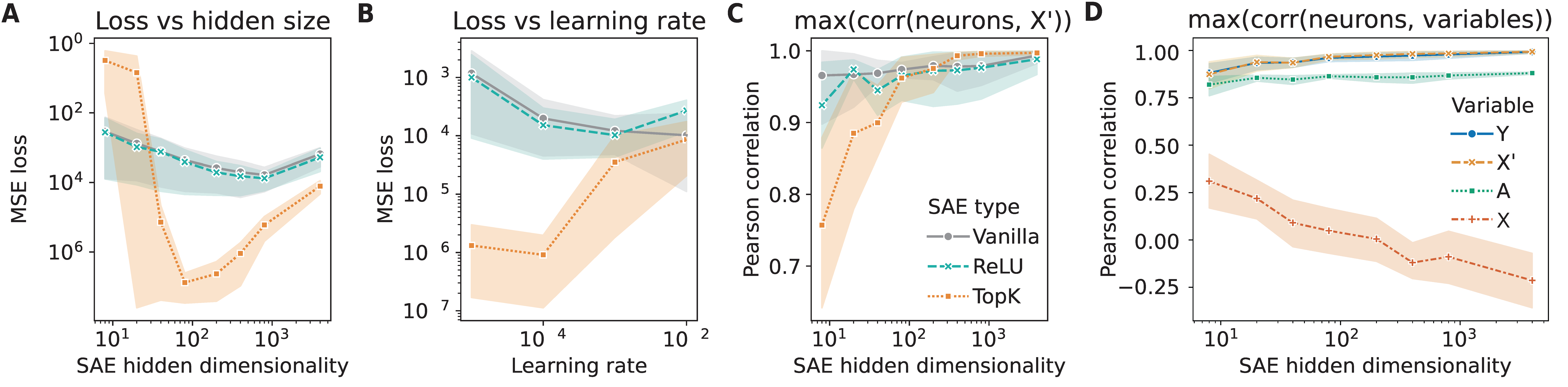}
\caption{\textbf{Performances of different SAE architectures on the small simulation data.} Performances of the three SAE types are presented as line plots with points depicting the average values over hyperparameter runs per SAE type ($N$ listed with each plot) and lines and areas as projections of mean and $95\,\%$ confidence, respectively. Vanilla, ReLU, and TopK SAEs are identified in legend C. \textbf{A} MSE loss against hidden dimensionality (learning rate $10^{-4}$, $N=5$). \textbf{B} MSE loss against learning rates ($N=40$). \textbf{C} Maximum Pearson correlation between SAE neurons and hidden variable $X'$ of the simulated data against hidden dimensionality (learning rate $10^{-4}$, $N=5$). \textbf{D} Recovery of simulation variables. Maximum Pearson correlation between SAE neurons and hidden variables of the simulated data against hidden dimensionality. Variables are explained in the legend to the right (learning rate $10^{-4}$, $N=16$ samples per point including all SAE types).
}
\label{fig:sim_metrics}
\end{figure}

\begin{figure*}[h]
\centering
\includegraphics[width=1\linewidth]{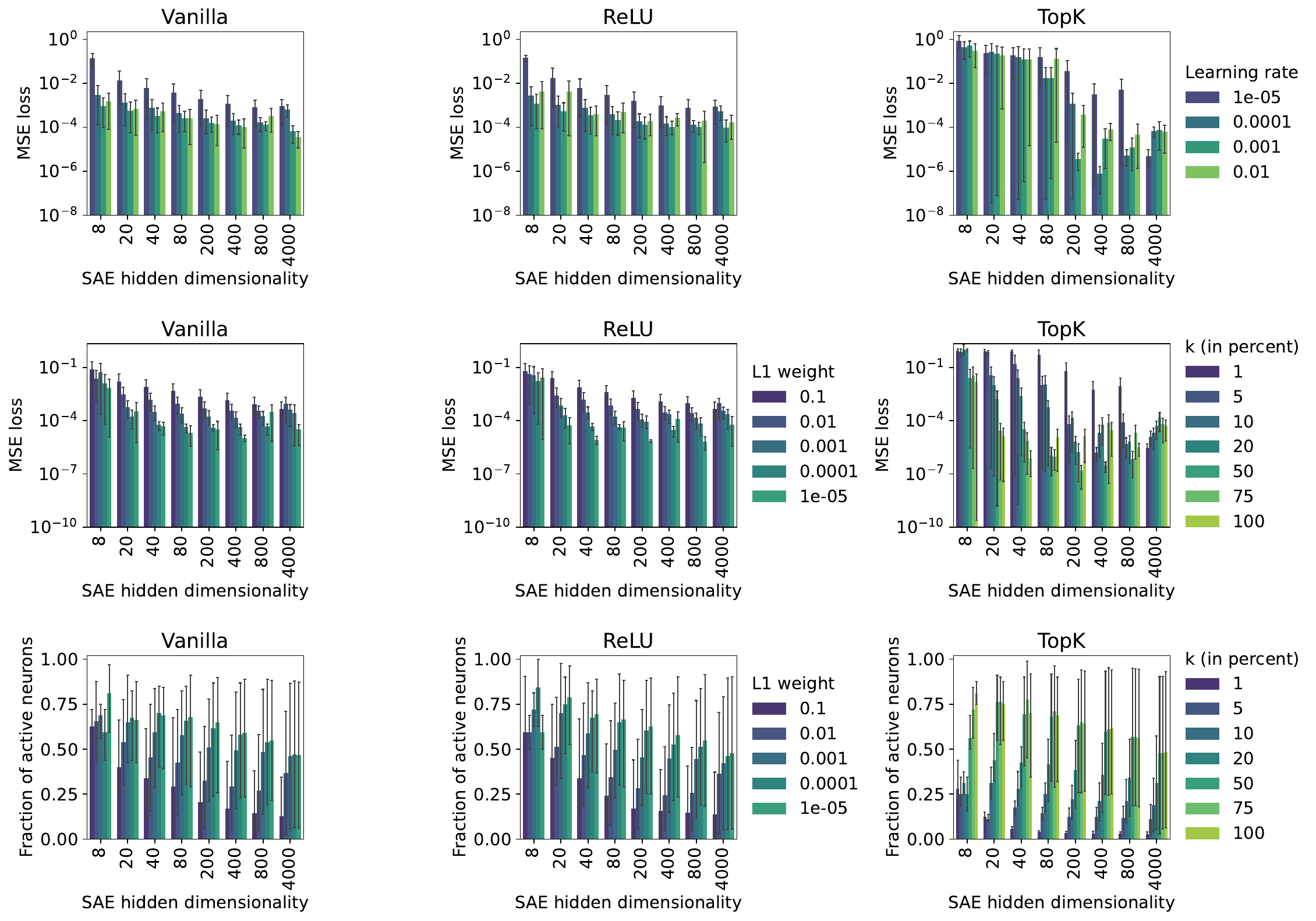}
\caption{\textbf{Hyperparameter bar plots of different types of SAEs trained on representations from the simulation experiment (latent dimension 4).} Columns depict performances for the SAE types Vanilla, ReLU, and TopK. Rows present different combinations of performance metrics. \textbf{A)} MSE loss against the hidden dimensionality colored by learning rate. $N=5$ runs per bar. \textbf{B)} Same as A colored by the sparsity penalty (L$_1$ weight for Vanilla and ReLU, $k$ in percent of hidden units for TopK). $N=4$ runs per bar.\textbf{C)} Fraction of active neurons against the hidden dimensionality colored by the sparsity penalty. $N=4$ runs per bar. Error bars indicate the $95$th confidence interval.}
\label{fig:supp_sae_performance}
\end{figure*}

\begin{figure*}[h]
\centering
\includegraphics[width=1\linewidth]{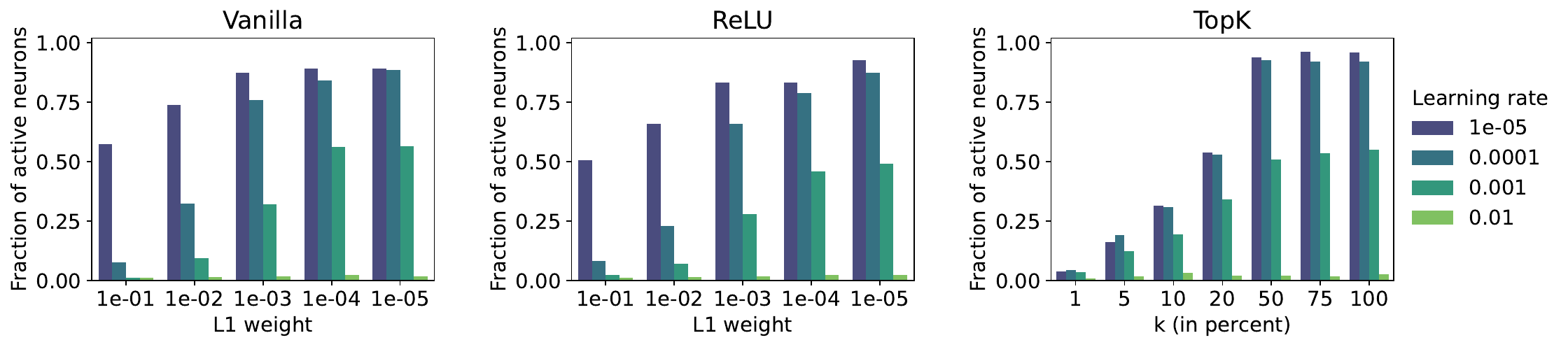}
\caption{\textbf{Influence of learning rate on the number of active neurons.} Bar plots of the three SAE types trained on the same representations as above for a hidden dimension of 400 ($100 \times \mathrm{latent}$). Columns depict performances for the SAE types Vanilla, ReLU, and TopK. The fraction of active neurons is plotted against the sparsity penalty colored by the learning rate ($N=1$).}
\label{fig:supp_sae_lr}
\end{figure*}

\begin{figure*}[h]
\centering
\includegraphics[width=1\linewidth]{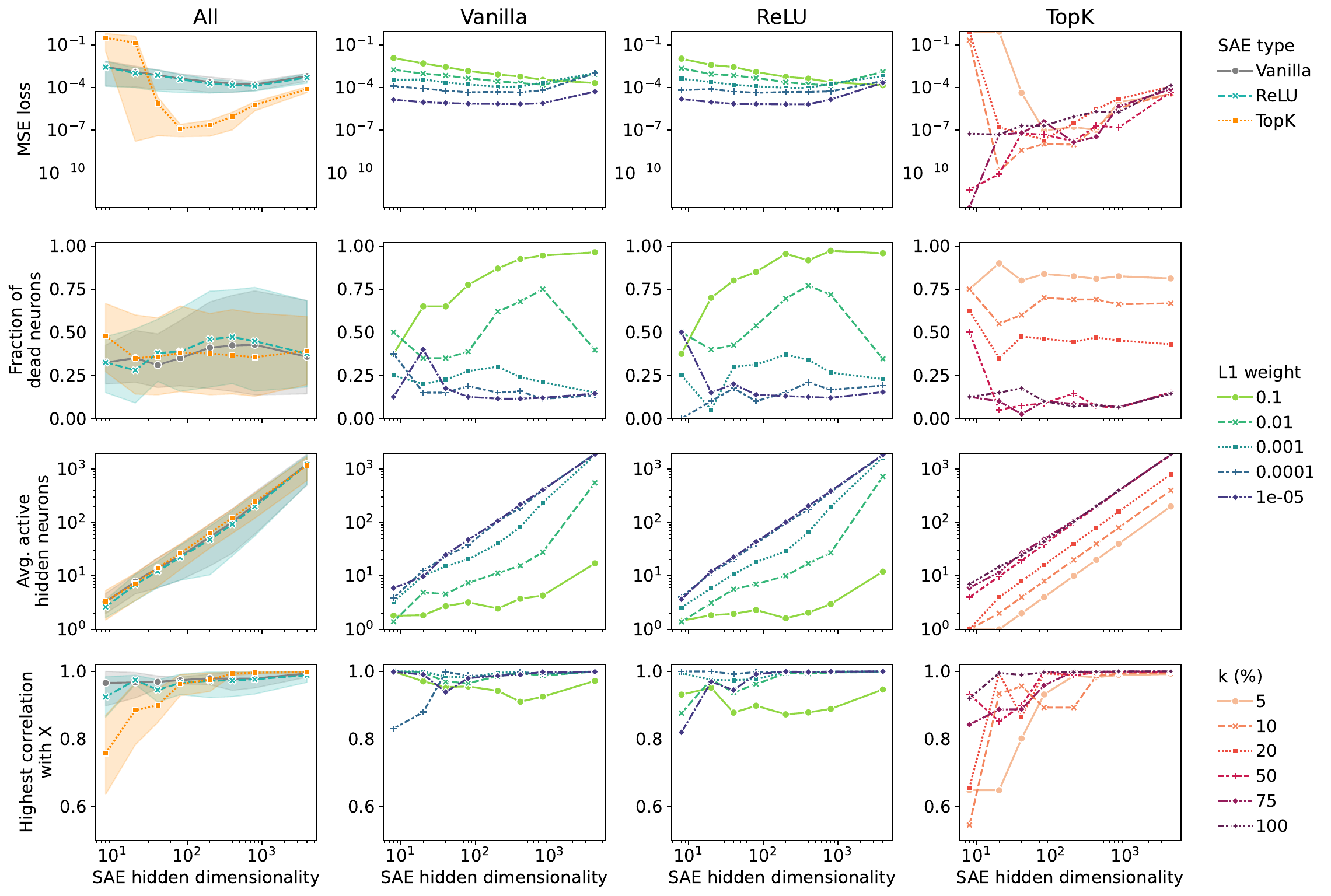}
\caption{\textbf{Performance comparison of SAEs for learning rate $10^{-4}$.} The first column presents accumulated line plots of specific metrics for the three different model types over the hidden dimensionality ($N=5$ and $N=6$ samples per point for Vanilla/ReLU and TopK, respectively) with the area as the $95$th confidence interval. The other three columns show the individual data points as line plots colored by the sparsity penalty. Legends to the right. The rows depict different metrics on the y axes: MSE loss, fraction of dead neurons, average number of firing neurons per sample, highest Pearson correlation of SAE neurons with variables $X$.}
\label{fig:supp_sae_performance_lr4}
\end{figure*}

\begin{figure*}[h]
\centering
\includegraphics[width=1\linewidth]{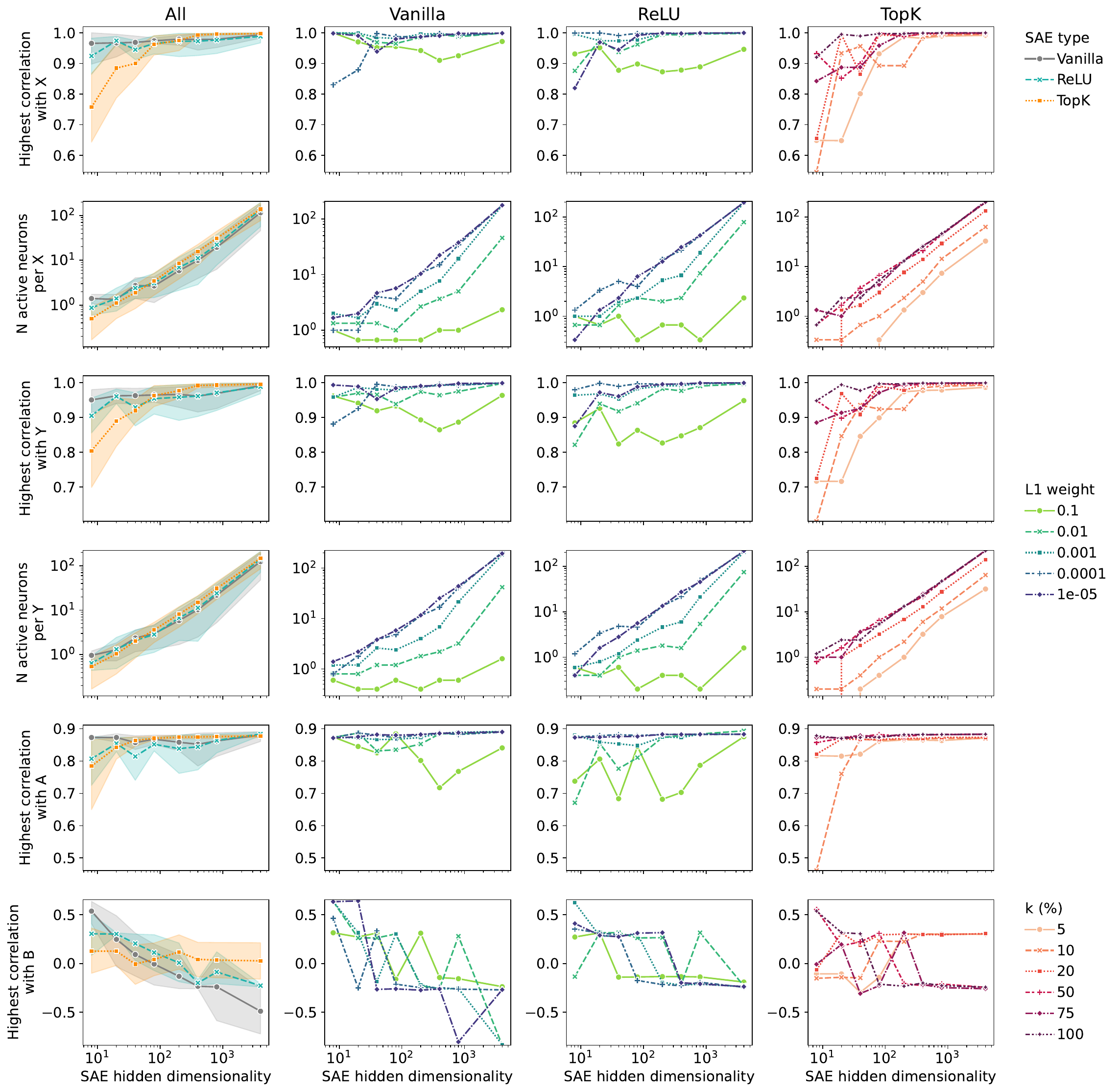}
\caption{\textbf{Comparison of variable recovery in different SAEs for learning rate $10^{-4}$.} Same as Supplementary Figure \ref{fig:supp_sae_performance_lr4} with different metrics on the y axes. Metrics refer to the highest Pearson correlation of SAE neurons with the simulation variables, as well as the number of corresponding SAE neurons with a correlation threshold of $>95\,\%$.}
\label{fig:supp_sae_recovery}
\end{figure*}

\begin{figure*}[h]
\centering
\includegraphics[width=1\linewidth]{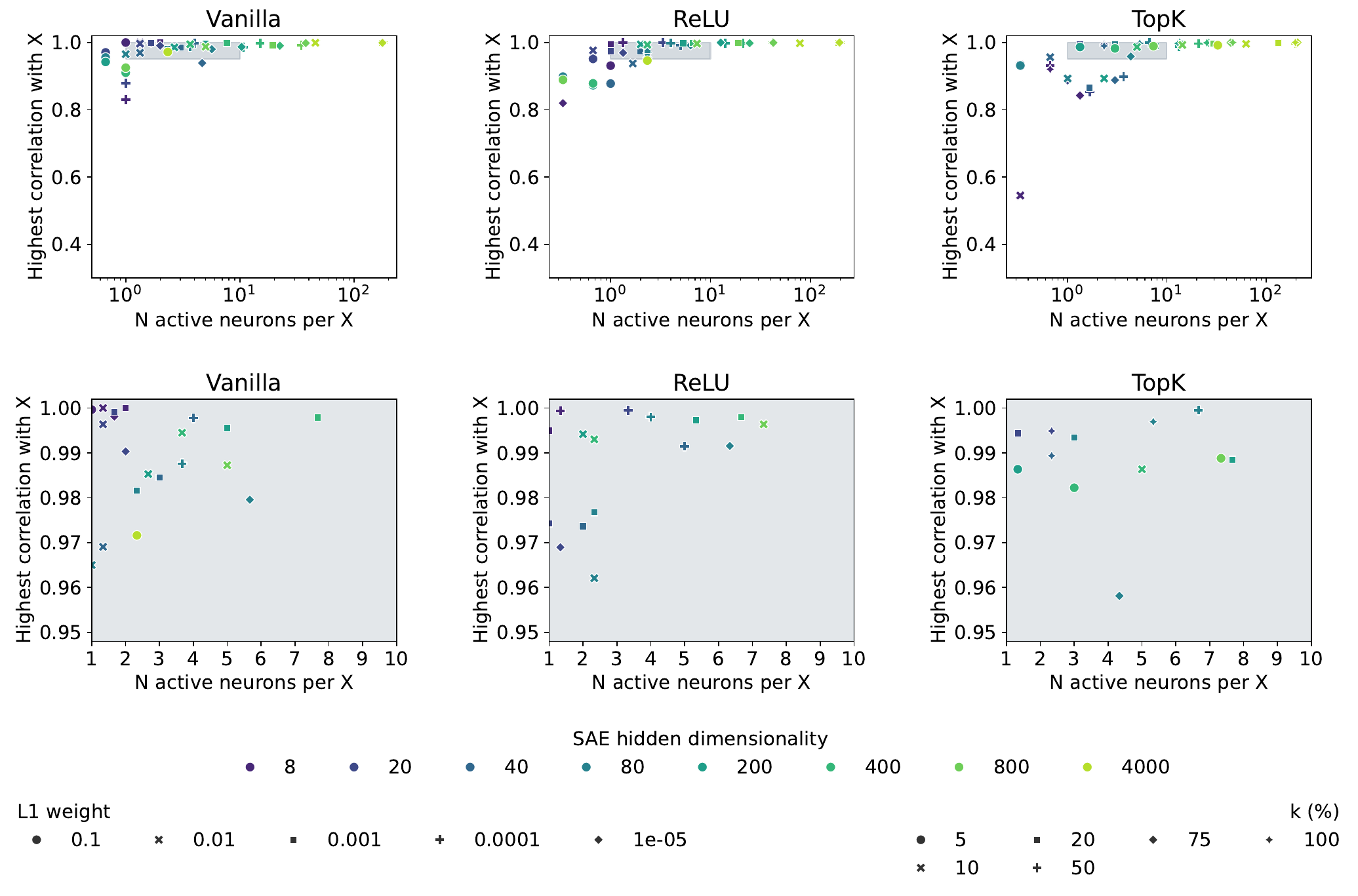}
\caption{\textbf{Sensitivity and specificity of SAE neurons for variable $X$.} Highest correlations of SAE neurons plotted against the number of active SAE neurons with a correlation threshold of $>95\,\%$ for Vanilla, ReLU, and TopK SAEs (columns from left to right). Colors indicate the hidden dimensionality. Data point styles indicate the sparsity penalty, explained in the legend at the bottom. The top row shows all model setups. The bottom row depicts the area highlighted as a grey box in the top row.}
\label{fig:supp_sae_recovery_x}
\end{figure*}

\begin{figure*}[h]
\centering
\includegraphics[width=1\linewidth]{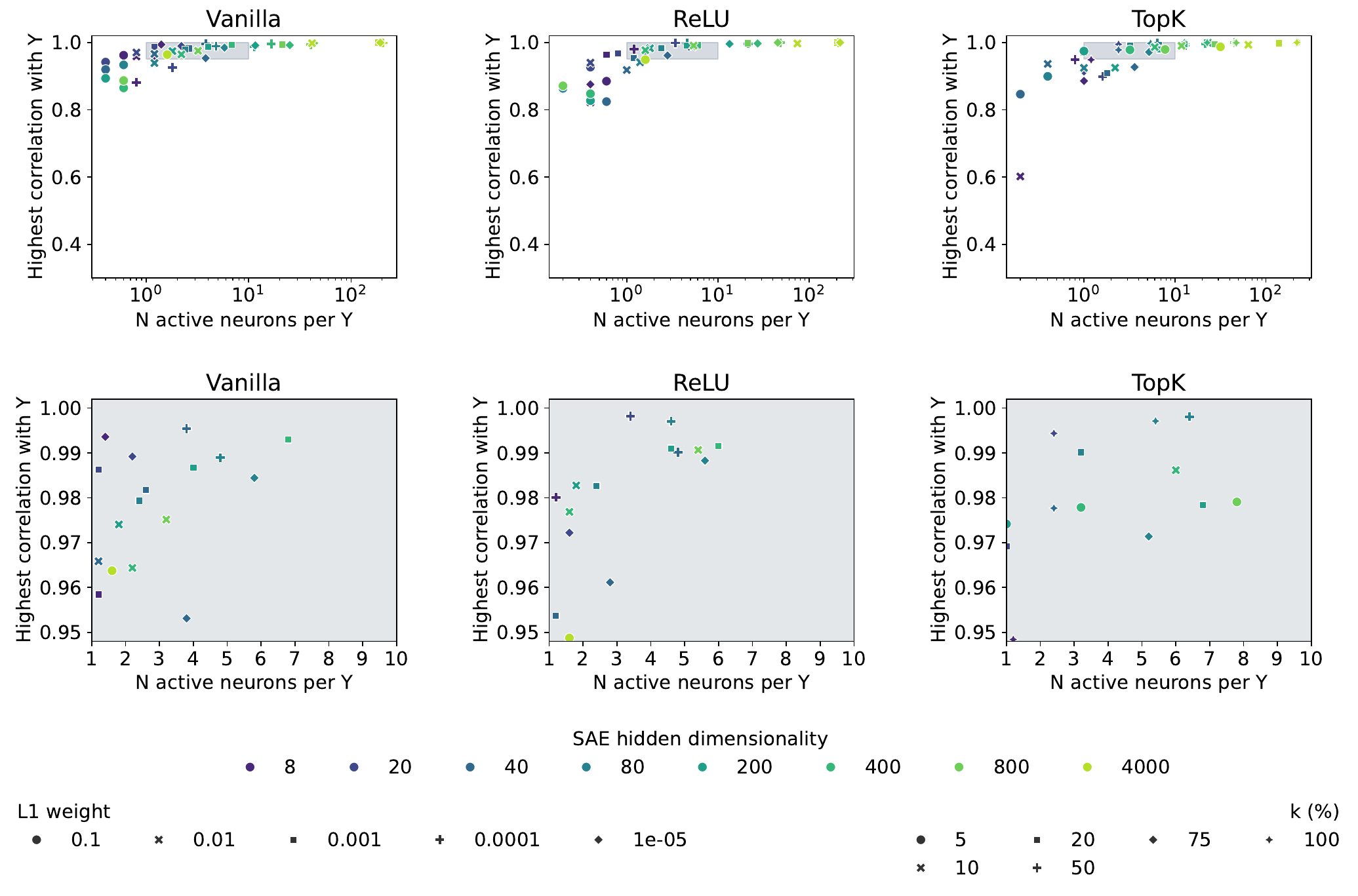}
\caption{\textbf{Sensitivity and specificity of SAE neurons for variable $Y$.} Highest correlations of SAE neurons plotted against the number of active SAE neurons with a correlation threshold of $>95\,\%$ for Vanilla, ReLU, and TopK SAEs (columns from left to right). Colors indicate the hidden dimensionality. Data point styles indicate the sparsity penalty, explained in the legend at the bottom. The top row shows all model setups. The bottom row depicts the area highlighted as a grey box in the top row.}
\label{fig:supp_sae_recovery_y}
\end{figure*}

\begin{figure*}[h]
\centering
\includegraphics[width=1\linewidth]{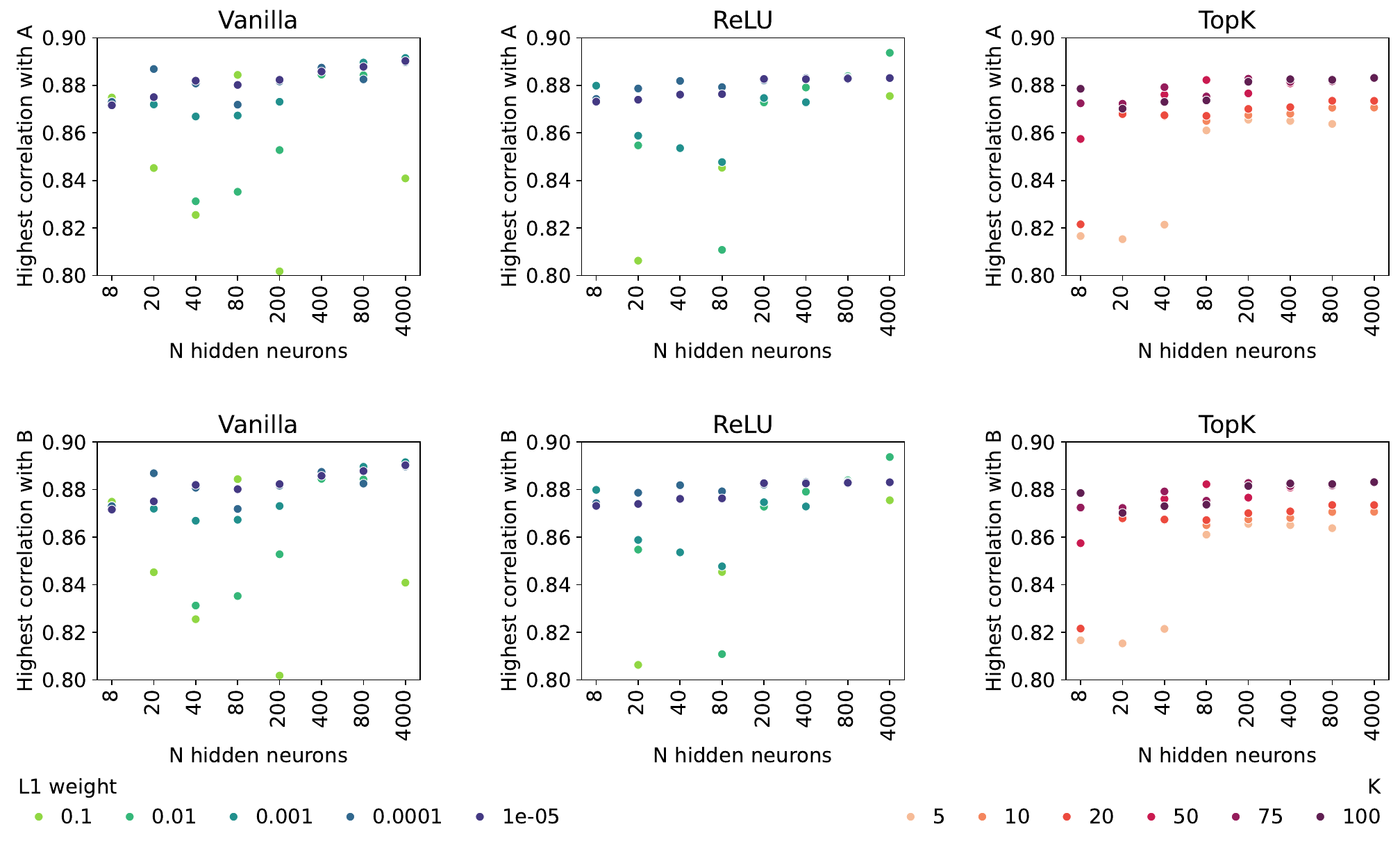}
\caption{\textbf{Sensitivity and specificity of SAE neurons for variables $A$ (top) and $B$ (bottom).} Highest correlations of SAE neurons plotted against the dimensionality of the SAE hidden space for Vanilla, ReLU, and TopK SAEs (columns from left to right). Colors indicate the sparsity penalty, explained in the legend at the bottom. The top row shows all model setups.}
\label{fig:supp_sae_recovery_ab}
\end{figure*}


\begin{figure}[h]
\centering
\includegraphics[width=0.35\linewidth]{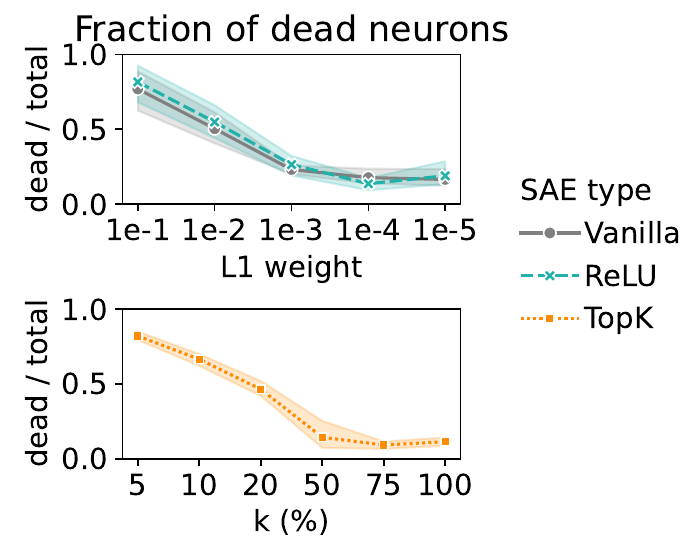}
\includegraphics[width=0.6\linewidth]{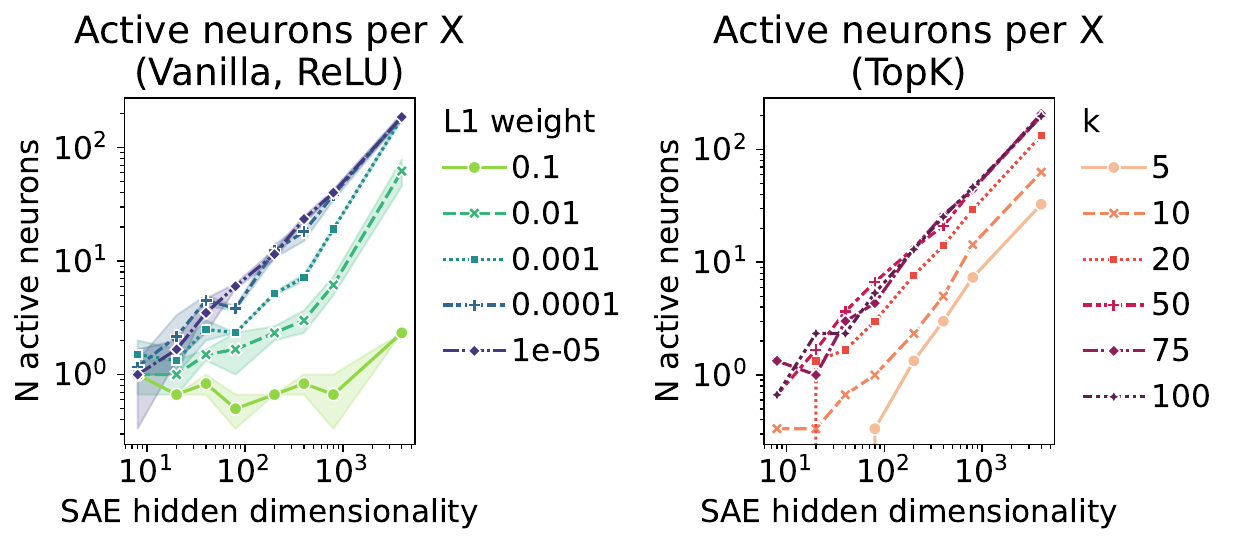}
\caption{\textbf{Redundancy of SAE features.} The two line plots show the number of active neurons per variable $X$ colored by sparsity parameter for Vanilla/ReLU (sparsity parameter: $\mathrm{L1}$ weight) and TopK (sparsity parameter: $k$) SAEs, respectively. The number of features are plotted against the total number of hidden neurons in the SAE. Line plots are set up as in Figure \ref{fig:sim_metrics} with $N=2$ and $N=1$ samples per point, respectively.}
\label{fig:sim_scaling}
\end{figure}

\begin{figure}[h]
\centering
\includegraphics[width=0.9\linewidth]{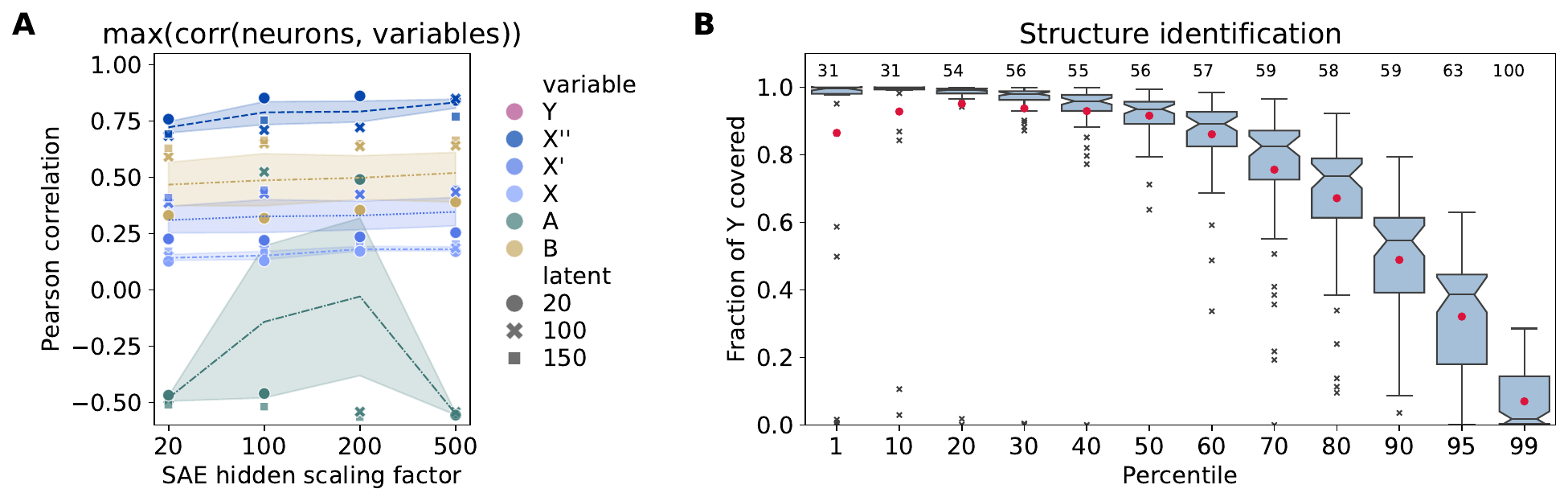}
\caption{\textbf{Recovery of large simulation variables and structure in SAE features. Left:} Maximum Pearson correlation between SAE neurons and hidden variables of the simulated data against hidden scaling factor ($N=3$). Points are colored by variable and the style depicts the AE latent dimensionality (legend on the right). \textbf{Right:} Boxplot of the fraction of ``genes'' $Y$ regulated by individual $X''$ variables connected to best matching SAE features. The x axis presents percentiles of the cosine similarities between SAE features and $Y$. The boxplot center line depicts the median, notches the $95\,\%$ confidence interval, and error bars $1.5$ times the interquartile range. Red dots present the means and numbers above indicate the number of samples per boxplot (= the number of $X$ variables out of $100$ that were matched with an SAE feature).}
\label{fig:sim_recovery_a}
\end{figure}

\begin{figure*}[h]
\centering
\includegraphics[width=1\linewidth]{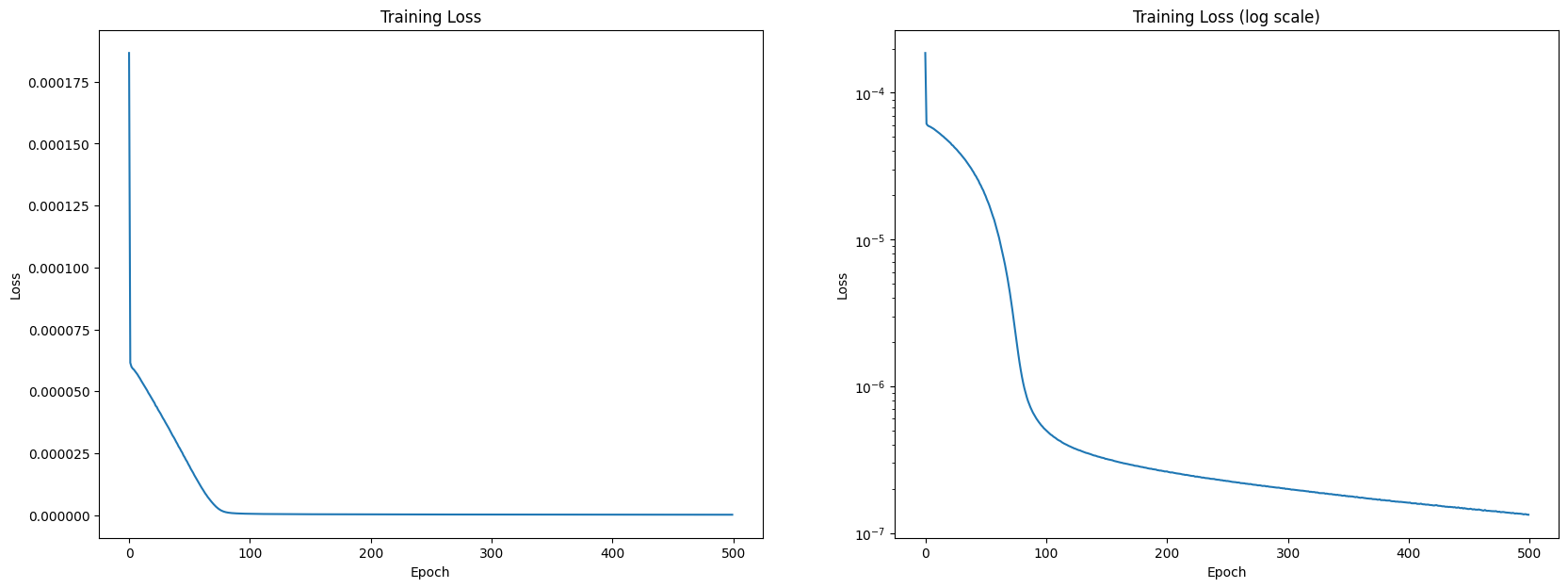}
\caption{\textbf{SAE training loss curve for the human bone marrow model.} The reconstruction loss (MSE) is plotted against the epochs. The right plot depicts the log-scaled loss.}
\label{fig:supp_sc_loss}
\end{figure*}

\begin{figure*}[h]
\centering
\includegraphics[width=1\linewidth]{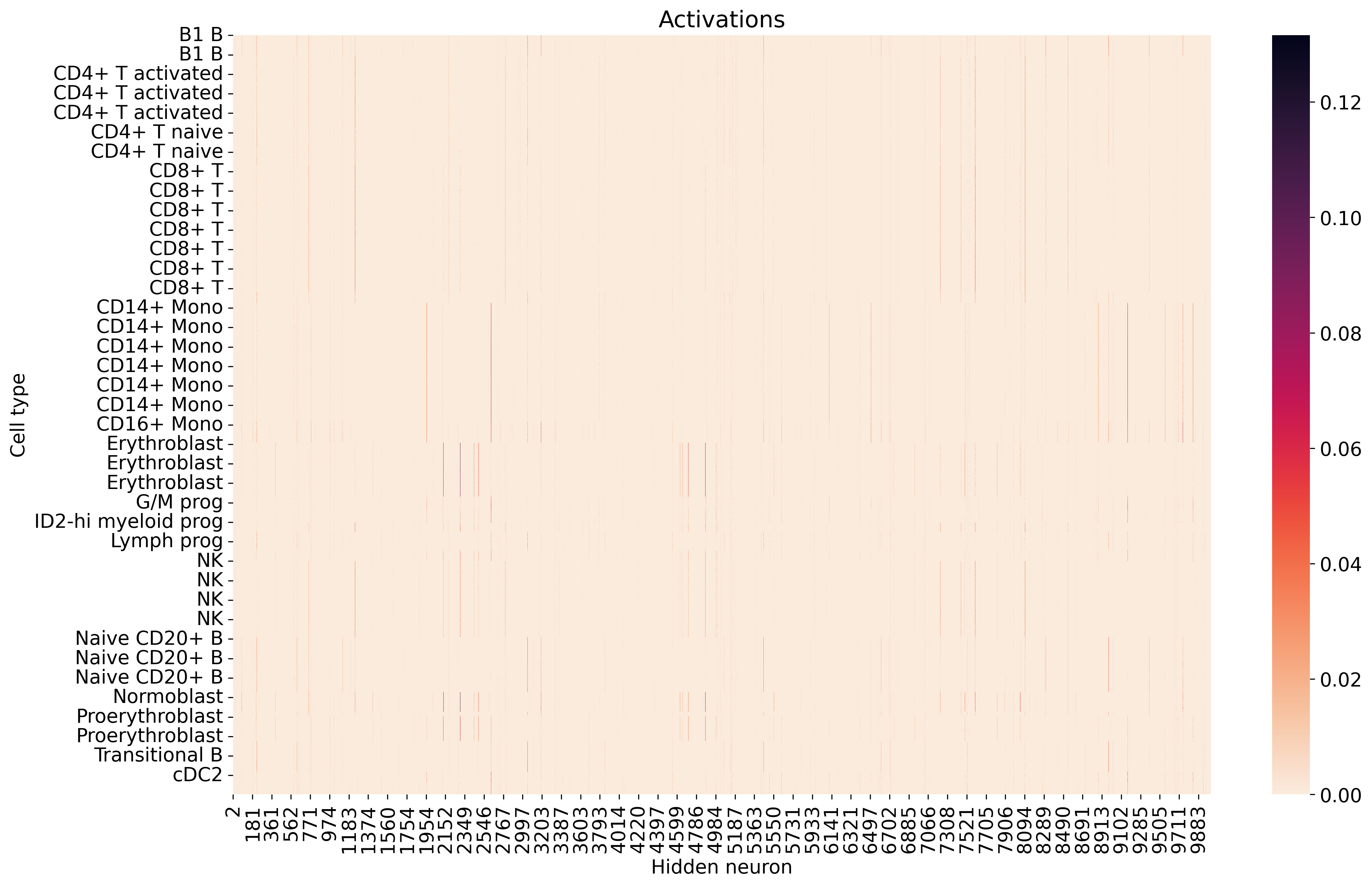}
\caption{\textbf{Heatmap of SAE activations from human bone marrow.} All samples are sorted by cell type on the y axis. All activations of active neurons are plotted on the x axis. The legend on the right describes the color range of the activations.}
\label{fig:supp_sc_activ}
\end{figure*}

\begin{figure*}[h]
\centering
\includegraphics[width=0.9\linewidth]{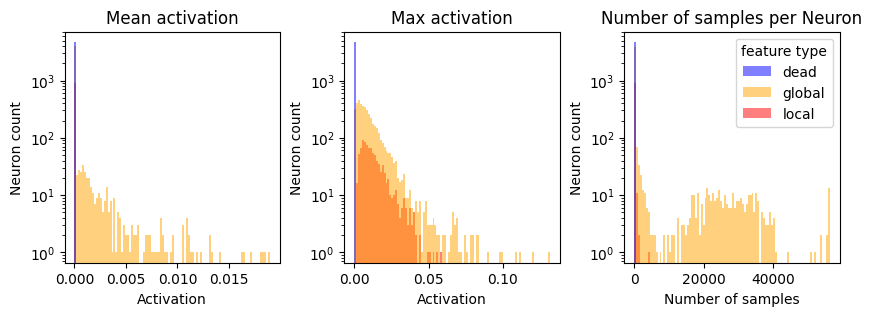}
\caption{\textbf{Feature activations of the SAE trained on human bone marrow single-cell data.} Log-scale neuron counts are plotted against mean activation, maximum activation, and the number of samples per neuron. Histograms are colored by the type of neuron.}
\label{fig:supp_sc_local_global}
\end{figure*}

\begin{figure*}[h]
\centering
\includegraphics[width=0.9\linewidth]{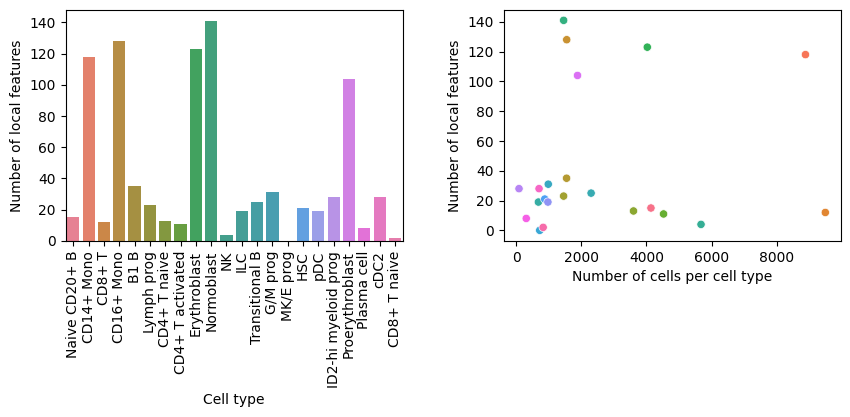}
\caption{\textbf{Distribution of local features among cell types.} The left shows a bar plot of the number of local features associated with each cell type. The right shows the number of local features plotted against the number of cells per cell type. Colors are the same as on the left.}
\label{fig:supp_sc_local}
\end{figure*}

\begin{figure*}[h]
\centering
\includegraphics[width=1\linewidth]{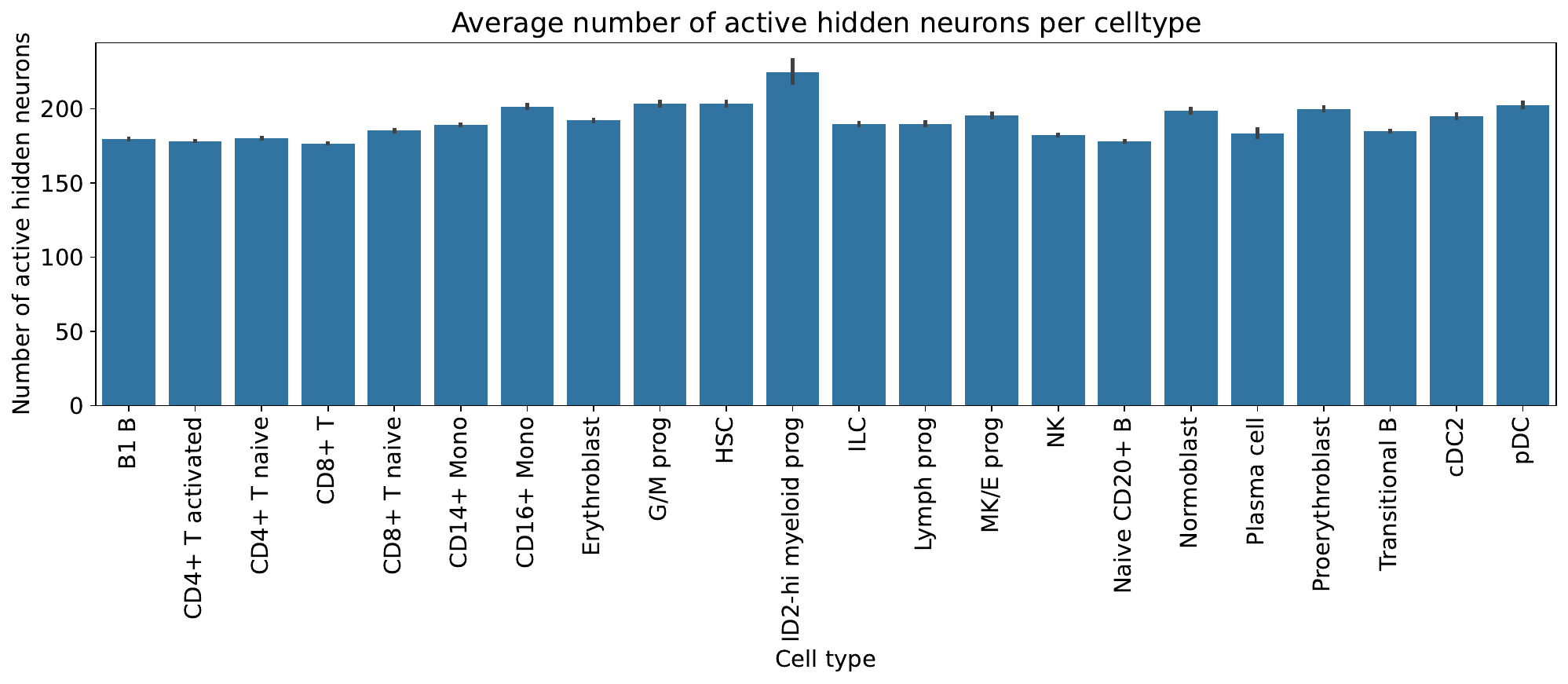}
\caption{\textbf{Average firing neurons per cell type.} Bar plots of the number of firing neurons per sample, plotted by cell type. Error bars indicate the $95$th confidence interval.}
\label{fig:supp_sc_avg_ct}
\end{figure*}

\begin{figure*}[h]
\centering
\includegraphics[width=1\linewidth]{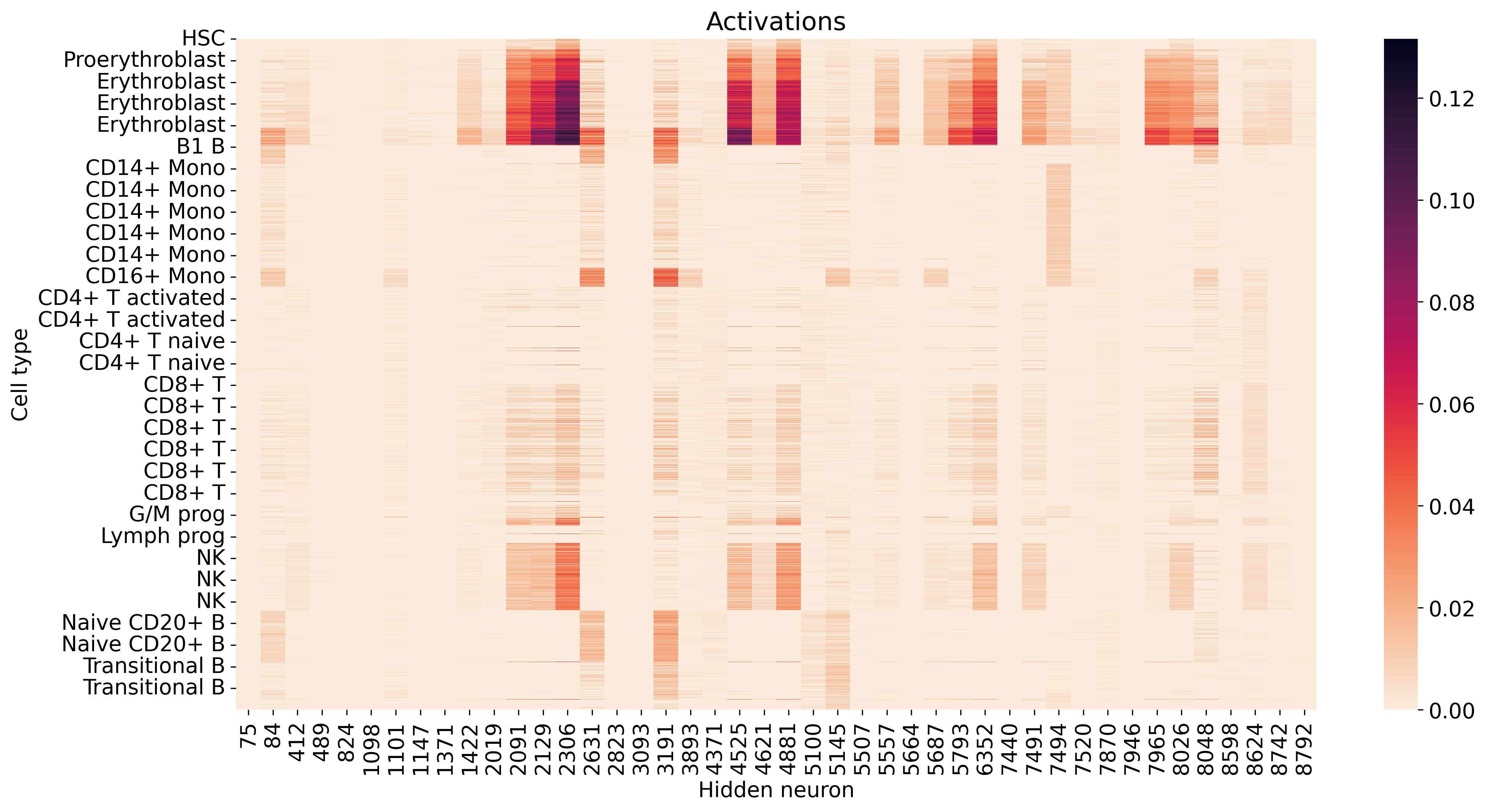}
\caption{\textbf{Activations of potential features for red blood cell development.} All samples are sorted by cell type on the y axis. Activations of neurons that fulfilled the requirements for red blood cell development are plotted on the x axis. The legend on the right describes the color range of the activations.}
\label{fig:supp_sc_activ_hemato}
\end{figure*}

\begin{figure*}[h]
\centering
\includegraphics[width=0.5\linewidth]{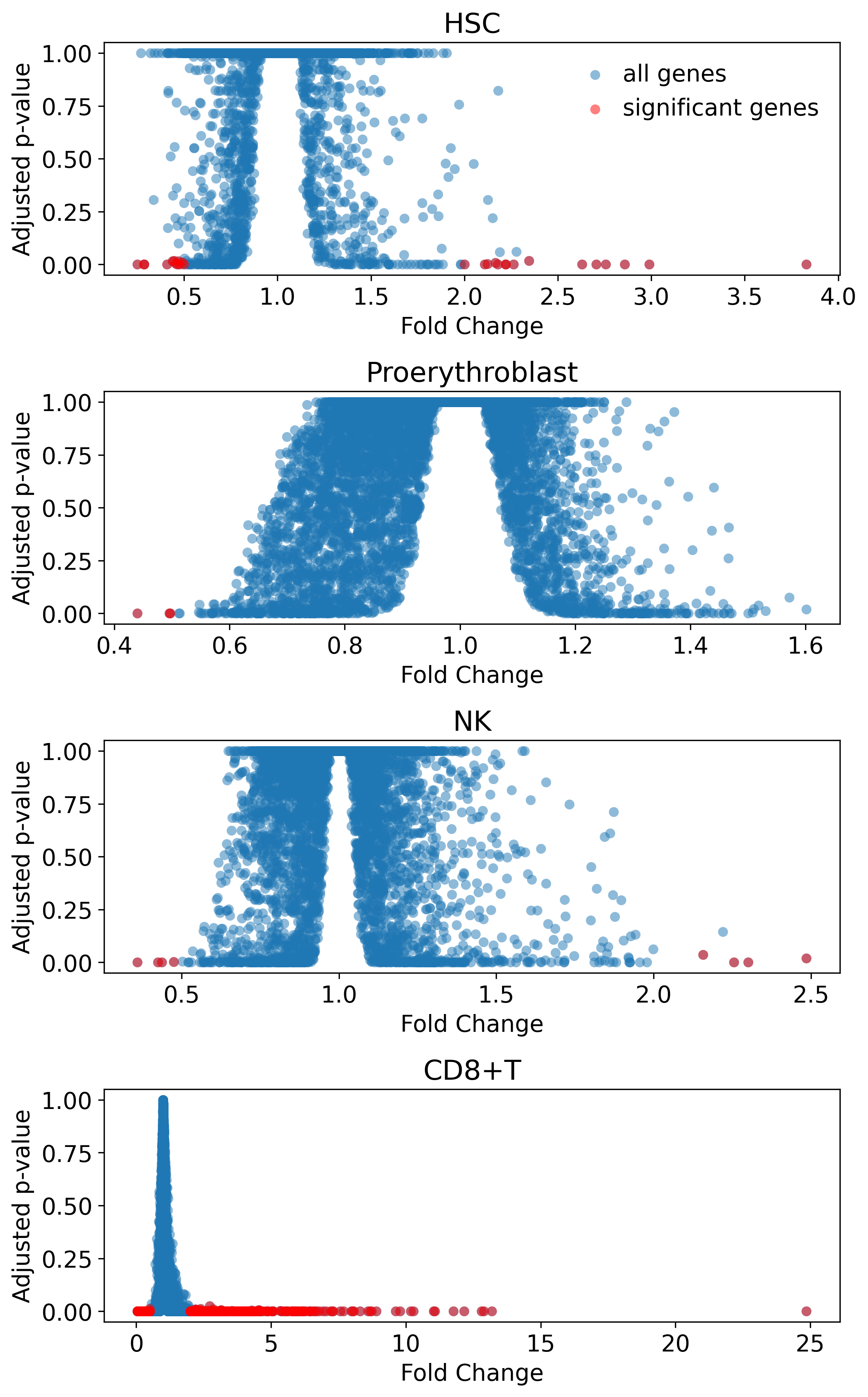}
\caption{\textbf{Differential gene expression analysis of perturbation experiments.} The plot shows the adjusted p-values against the fold change for all genes modeled by multiDGD \citep{schuster_multidgd_2023}. Each row shows the results of one of the four experiments indicated by the plot titles. Red data points depict genes with an adjusted p-value below $0.05$ and a fold change below $0.5$ or above $2$ (see legend in the top plot).}
\label{fig:supp_sc_deg}
\end{figure*}

\begin{figure}[h]
\centering
\includegraphics[width=0.9\linewidth]{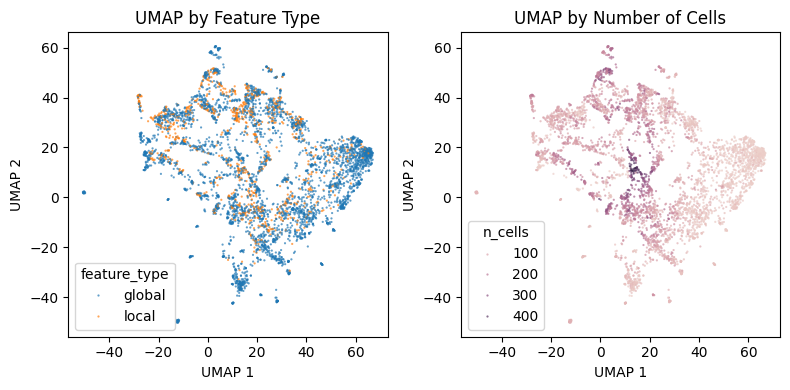}
\caption{\textbf{multiDGD SAE feature space UMAP.} The UMAP was computed with a minimum distance of 1, 10 neighbours, random seed 0, and a spread of 10. It is colored by feature type (left) and number of cells in which the feature is active (right).}
\label{fig:dgd_umap}
\end{figure}

\begin{figure}[h]
\centering
\includegraphics[width=0.9\linewidth]{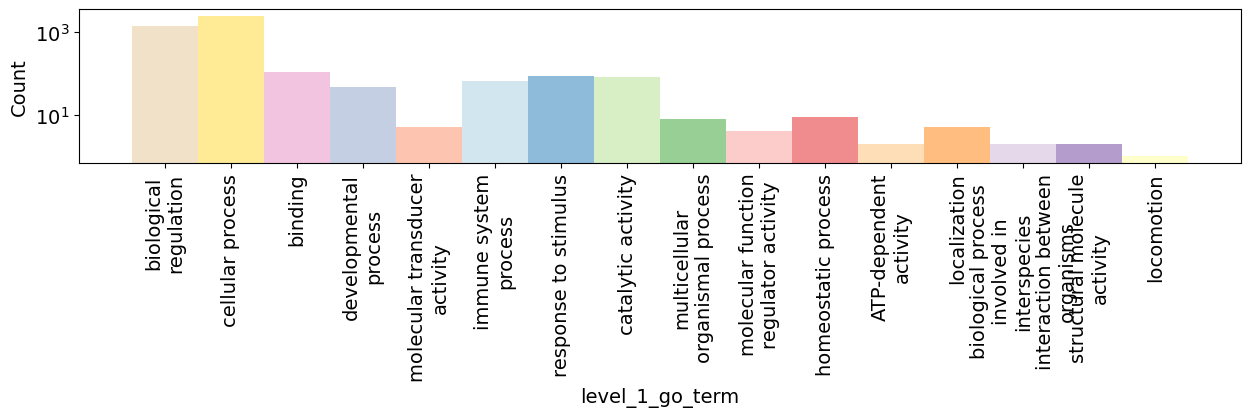}
\caption{\textbf{Frequency of individual GO terms.} The plot shows count histograms of all unique GO terms identified in the automated analysis colored by associated feature type (left) and GO term category (right).}
\label{fig:go_counts}
\end{figure}

\begin{figure}[h]
\centering
\includegraphics[width=0.5\linewidth]{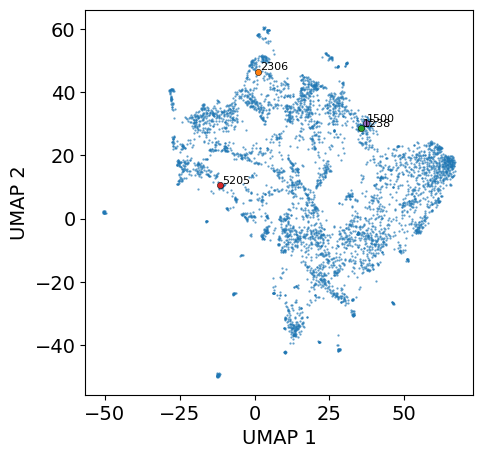}
\caption{\textbf{Single-cell SAE feature space UMAP indicating features from manual analysis.} Features 2306, 1238, 5205, and 1500 are highlighted by large colored dots and the feature id in black. All other features are depicted in blue.}
\label{fig:sim_feat_location}
\end{figure}

\begin{figure}[h]
\centering
\includegraphics[width=1\linewidth]{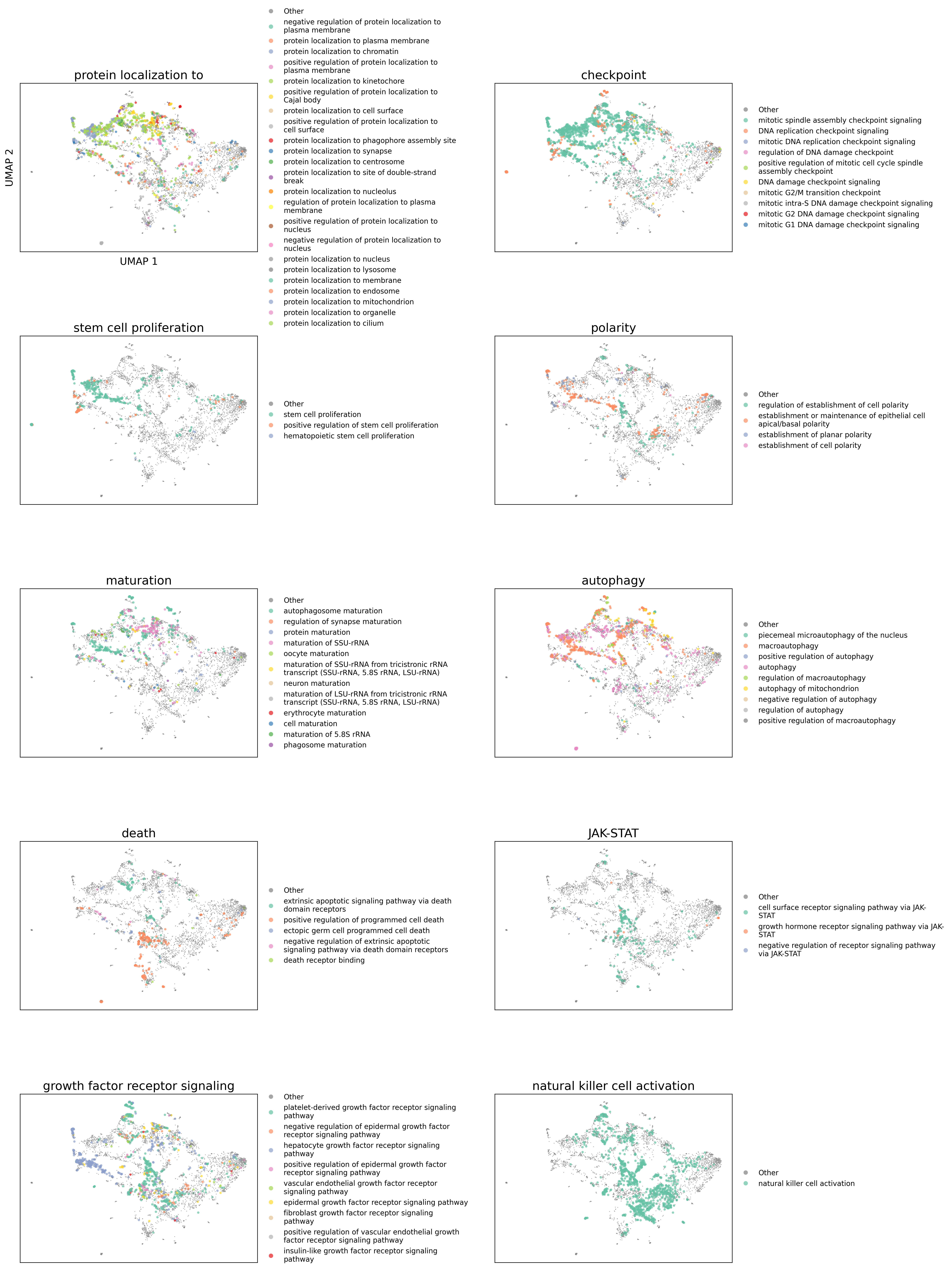}
\caption{\textbf{Probing multiDGD human bone marrow SAE feature space.} Words or concept snippets used for probing GO terms in the feature space are depicted in the title of each plot. Grey small background dots present all features (``other''). Colored, larger dots present all features in which the probing term was found. They are colored by the actual GO terms (legends to the right).}
\label{fig:sae_probing_supp}
\end{figure}

\begin{figure}[h]
\centering
\includegraphics[width=0.4\linewidth]{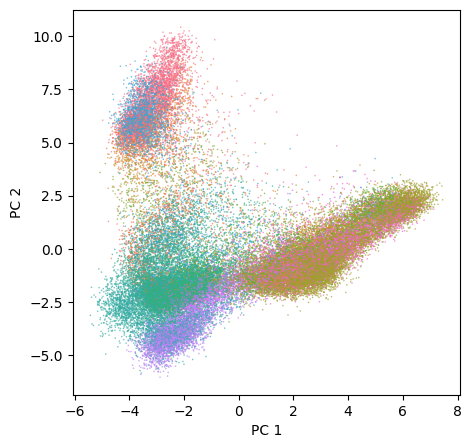}\includegraphics[width=0.52\linewidth]{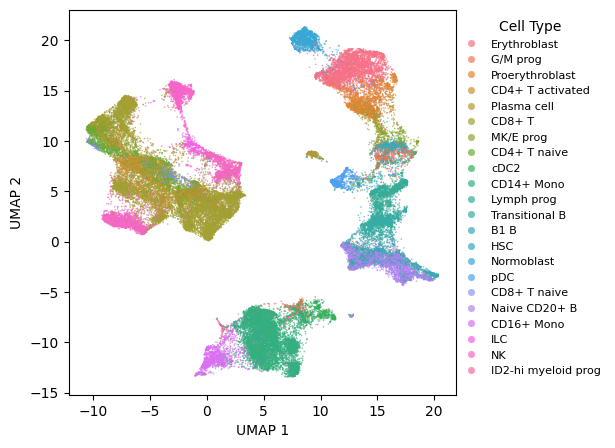}
\caption{\textbf{Geneformer embeddings of the human bone marrow data}. PCA on the left, colored by cell type (legend to the right). The right plot shows a UMAP with minimum distance 0.2, 20 neightbors, a spread of 0, and random seed 0.}
\label{fig:sae_geneformer_embedding}
\end{figure}

\begin{figure}[h]
\centering
\includegraphics[width=1\linewidth]{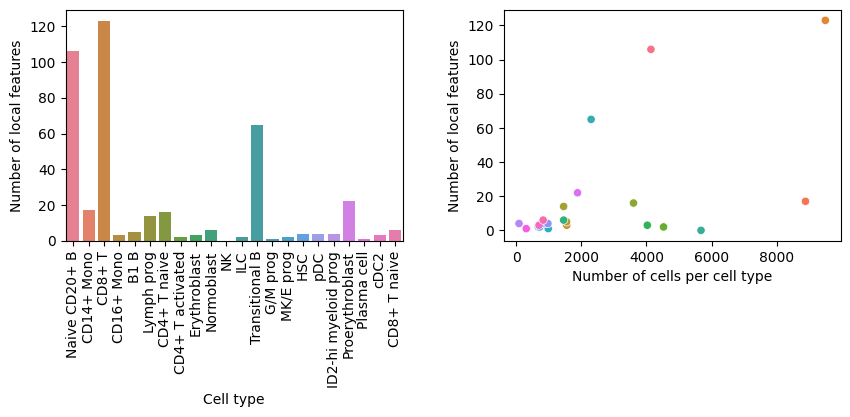}
\caption{\textbf{Distribution of local features among cell types in Geneformer embedding SAE.} The left shows a bar plot of the number of local features associated with each cell type. The right shows the number of local features plotted against the number of cells per cell type. Colors are the same as on the left.}
\label{fig:sae_geneformer_ct_feature_counts}
\end{figure}

\begin{figure}[h]
\centering
\includegraphics[width=0.6\linewidth]{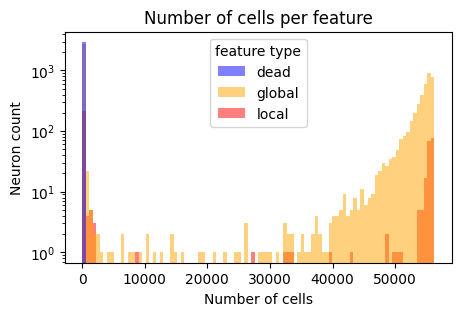}
\caption{\textbf{Histogram of features over cells of the Geneformer SAE trained on human bone marrow single-cell data.}}
\label{fig:sae_geneformer_cells_per_feature}
\end{figure}

\begin{figure}[h]
\centering
\includegraphics[width=0.9\linewidth]{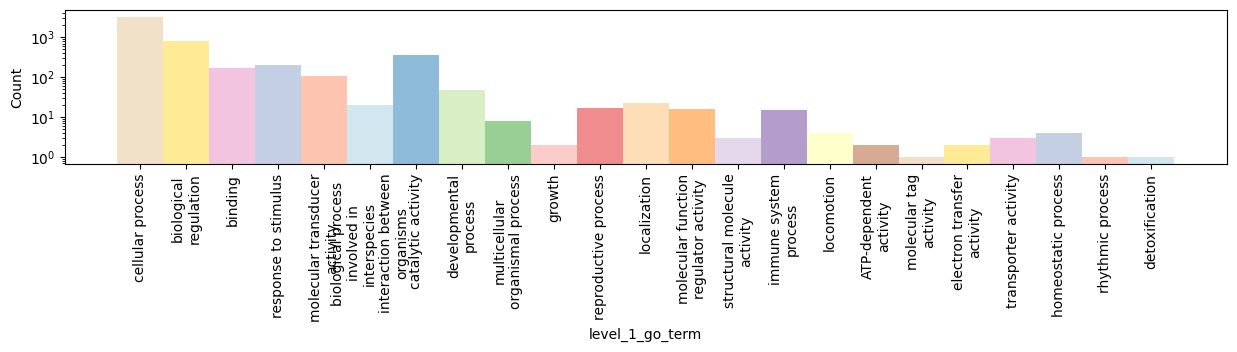}
\caption{\textbf{Frequency of individual GO terms.} The plot shows count histograms of all unique GO terms identified in the automated analysis colored by associated feature type (left) and GO term category (right).}
\label{fig:go_counts_geneformer}
\end{figure}

\begin{figure}[h]
\centering
\includegraphics[width=0.9\linewidth]{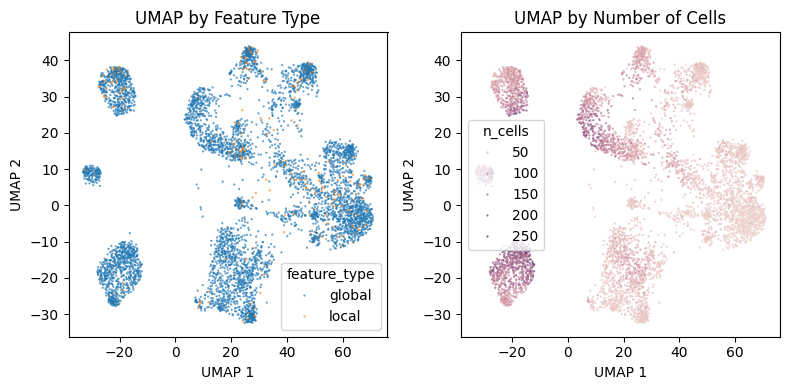}
\caption{\textbf{Geneformer SAE feature space UMAP.} The UMAP is colored by feature type (left) and number of cells in which the feature is active (right).}
\label{fig:geneformer_umap}
\end{figure}

\begin{figure}[h]
\centering
\includegraphics[width=1\linewidth]{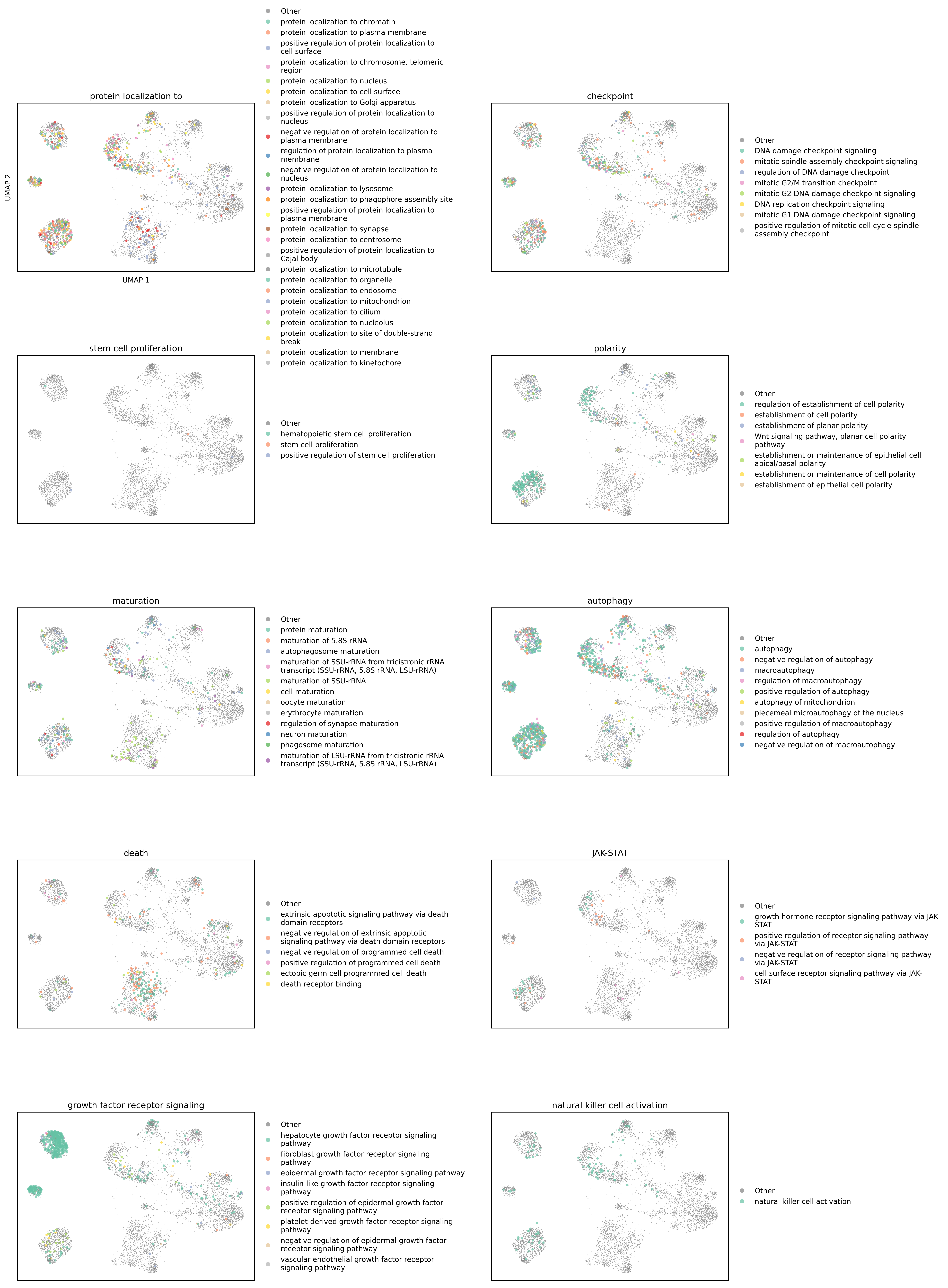}
\caption{\textbf{Probing Geneformer human bone marrow SAE feature space.} Words or concept snippets used for probing GO terms in the feature space are depicted in the title of each plot. Grey small background dots present all features (``other''). Colored, larger dots present all features in which the probing term was found. They are colored by the actual GO terms (legends to the right).}
\label{fig:sae_geneformer_probing}
\end{figure}

\begin{figure}[h]
\centering
\includegraphics[width=0.5\linewidth]{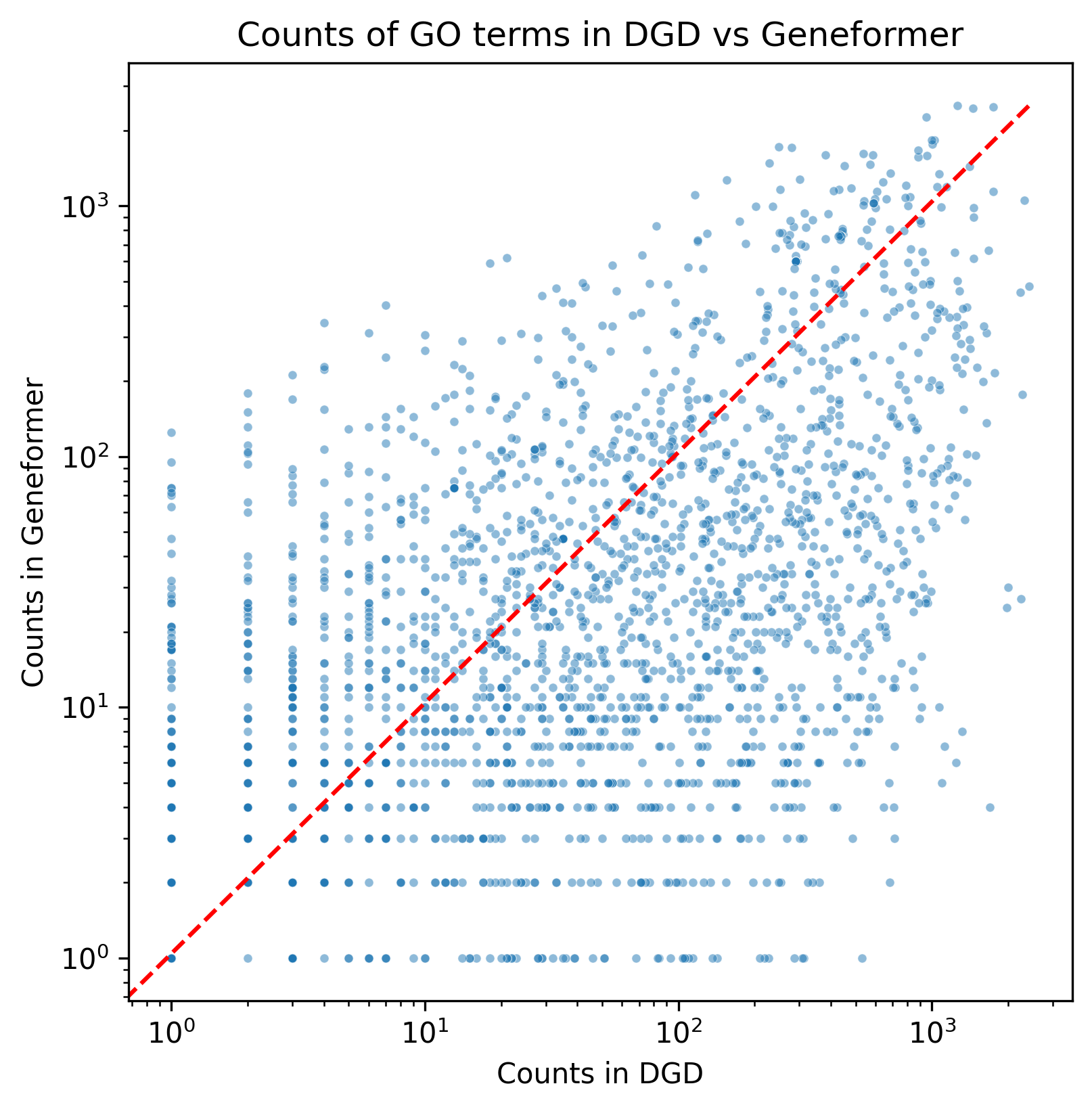}
\caption{\textbf{Value counts in DGD and Geneformer SAE feature space per shared GO term.}}
\label{fig:sae_geneformer_dgd_go_counts}
\end{figure}

\end{document}